\documentclass{article}
\usepackage[preprint]{tmlr}
\usepackage{amsmath,booktabs}
\usepackage{colortbl}
\definecolor{carvetint}{RGB}{232,241,248}
\usepackage{graphicx,xcolor}
\usepackage{fontawesome5}
\usepackage{subcaption}
\usepackage{microtype}
\usepackage{hyperref}
\usepackage{url}
\hypersetup{colorlinks=true,linkcolor=blue,citecolor=blue,urlcolor=blue,
  pdftitle={Model Carving: Training Transformers That Optimize Their Own Topology},
  pdfauthor={Kaden Zheng, Pratyusha Sharma}}

\newcommand{\carving}{\textsc{Model Carving}}

\begin{document}

\begin{center}
  {\fontfamily{phv}\selectfont\LARGE\bfseries Model Carving \faIcon{cut}:\\ Training Transformers That Optimize Their Own Topology \par}
  \vspace{1.5em}
  {\bfseries
        Kaden Zheng\textsuperscript{1},
    Pratyusha Sharma\textsuperscript{2}\par}
  \vspace{0.3em}
  \textsuperscript{1}Harvard University \qquad
  \textsuperscript{2}New York University
\end{center}

\begingroup                                             
\renewcommand{\thefootnote}{\fnsymbol{footnote}}
\footnotetext[1]{Project page: \url{https://kadenzheng.com/model-carving/}}
\endgroup                                 

\begin{abstract}
Machine learning has largely kept separate the choice of architecture and the process of training. A classic Transformer, for instance, trains a fixed set of attention and MLP sublayers at every layer, in the same order, and optimizes the weights within. 
This uniformity is a convention rather than a principled design choice, and it is likely suboptimal: the best architecture for a given task may be smaller, more efficient, and far from uniform.
We believe that, ideally, training itself should shape the model's topology. To this end, we introduce \carving{}, which learns both a Transformer’s weights and its architecture in a single run. By applying learnable, scalar gates to each sublayer, we allow the training loss to remove the model’s own components during training -- freeing a smaller heterogeneous network inside. We evaluate \carving{} by pre-training standard Transformers from 140M to 2.7B parameters and looped Transformers from 140M to 1.3B. At equal training compute, carved models reach a lower validation loss than dense models trained with up to 1.7 times more compute. Furthermore, carved models also outperform size-matched models produced by pruning or distillation, suggesting that the best way to obtain small models for edge devices is to carve them during training, not to compress a large one afterwards. This advantage both grows with scale and carries over to downstream tasks such as few-shot evaluation and generation. Because the removed sublayers no longer run, carved models also cost up to 34\% fewer FLOPs per token and decode up to 1.6 times faster without cost in quality. Our results show that architecture, like weights, should be learned and not fixed in advance, and that inside every large transformer is a smaller, faster, stronger one waiting to be carved out.

\end{abstract}

\section{Introduction}
\label{sec:introduction}

A long-standing north star for many in machine learning has been to build models as adaptive as the brain. One field that has emerged to pursue this has been \textit{adaptive computation}, which enables models to make their computations input-dependent at \textbf{inference-time} through techniques such as routing \citep{raposo2024mod,bae2025mor}, Mixture-of-Experts \citep{shazeer2017outrageously}, and early-exiting \citep{graves2016adaptive,schuster2022confident}. However, among the brain's most striking forms of plasticity is structural. During development, our brains significantly overproduce synapses and then gradually prune those whose metabolic cost outweighs their functional utility \citep{huttenlocher1979synaptic,chechik1998synaptic,bullmore2012economy}. In this sense, the architecture that performs the computation is shaped in part by its own use during\textbf{ training-time}.

Neural networks have largely kept architecture and training separate. Training optimizes a model's weights within an architecture fixed before the first gradient step is taken: a classic Transformer  \citep{vaswani2017attention} executes a homogeneous set of attention and MLP matrices at every layer, in the same fixed order. While this uniformity is a convention, it fails to leverage the rich expressivity of heterogeneous neural network structures. For instance, reordering sublayers alone can improve perplexity \citep{press2020sandwich}, whole layers of trained LLMs can be removed with modest degradation \citep{gromov2025ineffectiveness,men2024shortgpt}, and recent state-of-the-art models already break uniformity by hand, such as Gemma 3 interleaving local and global attention \citep{gemmateam2025gemma3} or Jamba interleaving attention and state-space layers \citep{lieber2024jamba}.

Where structure is learned, it is usually learned outside of training through \textit{neural architecture search} (NAS). Early methods trained thousands of candidate networks under a reinforcement-learning controller \citep{zoph2017neural} or an evolutionary loop \citep{real2019regularized,so2019evolved}. Weight-sharing methods amortized this cost by training a single supernet that contains every candidate operation, while optimizing a separate set of architecture parameters on held-out data~\citep{liu2019darts}. Fundamentally, the architecture is still chosen by a second optimization problem-- over a controller, a population, or architecture parameters-- run alongside, rather than within, the optimization of the model's weights.

\begin{figure}[t]
\centering
\includegraphics[width=\textwidth]{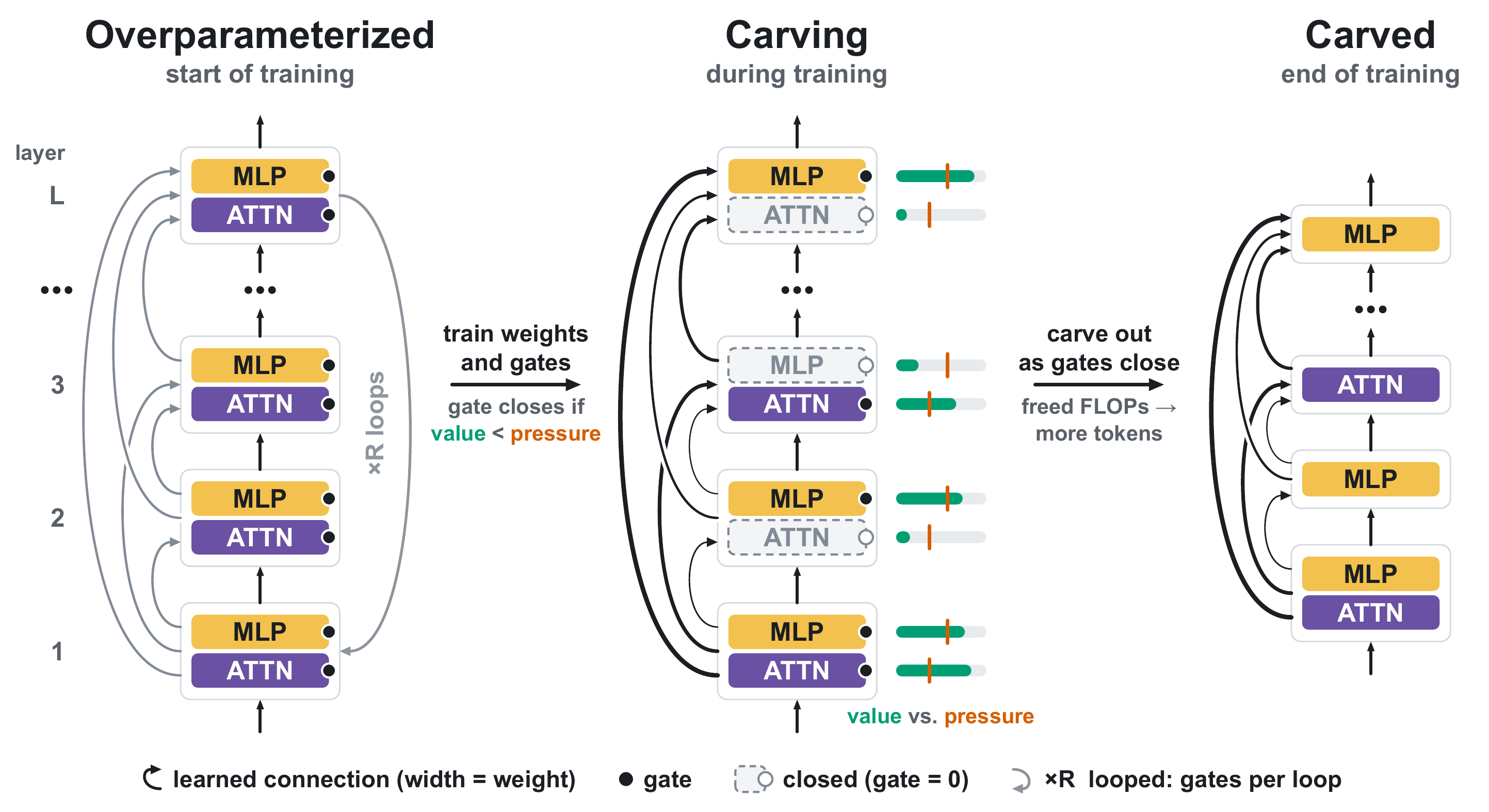}
\caption{\textbf{\carving{} trains the weights and the topology of a Transformer in one run.}
(Left) Training starts from an overparameterized Transformer: every sublayer writes through a gate (dot), and learned connections (arcs) join each pair of layers. A looped Transformer repeats the stack for $R$ loops, with its own gates in each loop.
(Middle) During training, a gate falls when its sublayer's contribution to the loss (green) is below its pressure (orange), and a gate at zero closes its sublayer.
(Right) The trainer removes the closed sublayers from the computation graph, and the freed FLOPs buy more training tokens, resulting in a smaller Transformer with a heterogeneous set of layers.}
\label{fig:method}
\end{figure}

In this paper, we present \carving{}, which collapses these into a single optimization problem. Rather than selecting an architecture and then training it, we take an overparameterized network and leverage \textit{gates} to allow the training loss to remove its own components-- freeing a smaller, heterogeneous network from inside (Figure~\ref{fig:method}). Each sublayer writes to the residual stream through a learned, scalar gate, and every sublayer that stays in the graph carries a \textit{pressure}. We construct this pressure online as the loss that that sublayer's compute would buy if spent as extra training tokens. When a sublayer contributes less to the loss than its pressure, its gate shrinks to zero and is subsequently removed from the computation graph. Therefore, by construction, a sublayer is kept only if its contribution to the loss is worth its compute. Gates and weights are optimized jointly, on the same objective, the same data, and in the same run. In this sense, we enable the architecture that performs the computation to be shaped in part by its own use.

We evaluate \carving{} on standard Transformers from 140M to 2.7B parameters and looped Transformers from 140M to 1.3B, comparing the resulting models at equal training compute with ``full-depth'' models from which \carving{} starts, and with ``shallower-depth'' models constructed at the carved models' inference cost. At equal training compute, \carving{} reaches a loss that full-depth models would need up to 1.7 times the compute to reach, matches or improves their downstream performance, and outscores public models trained with several times its compute, while running up to 34\% fewer inference FLOPs per token. Carved models also decode up to 1.61 times faster at each model's largest batch. We further find that \carving{}'s advantage over standard models grows with compute, namely a compute-optimal carved model reaches a lower loss than full-depth models in both families, and the gap widens with scale, which suggests that the convention of homogeneously repeated layers is not the most efficient design. 

\carving{} also pushes the frontier for building ML models for edge computing. Edge devices, where memory and inference compute are scarce, need models that are small and cheap to run, and the dominant paradigm of building these is to start from a larger model that is pruned and retrained \citep{han2015learning} or used as a teacher for model distillation \citep{hinton2015distilling}. 
\carving{} outperforms both pruning and distillation from larger models on the quality-efficiency frontier, and its lead is robust to weight quantization.

\section{\carving{}}
\label{sec:carving}

The central insight of \carving{} is that we can simultaneously train the weights of a model and allow it to learn its own architecture by leveraging \textit{gates}.
Given a specified training FLOPs budget, we incentivize the model to \textbf{remove the components whose contribution to the loss is less than their compute cost} by shrinking their gates to zero.
\carving{} has four parts, all within one training run: (1) initialize an overparameterized, gated Transformer that defines what can be removed (Section~\ref{sec:model}); (2) define a training objective that places a pressure on each sublayer for its compute (Section~\ref{sec:price}); (3) estimate that pressure online from the run's own learning curve (Section~\ref{sec:reading}); and (4) apply the pressure to the gates, which drives the gates of sublayers that contribute less than their pressure to zero, after which the trainer removes those sublayers from the computation graph (Section~\ref{sec:gate-update}). 

\subsection{Overparameterizing and gating a Transformer}
\label{sec:model}

An intuition for \carving{} may be obtained from biology, where synaptic pruning gradually eliminates weak synapses as the brain matures.
Likewise, we start \carving{} from an \textit{overparameterized} Transformer that defines what can be removed, in which every sublayer is present and additional residual connections join each pair of distinct layers.
\carving{} then removes the sublayers that do not earn their compute during training, while each connection grows or shrinks through its learned coefficient -- thereby ``carving'' an architecture from within.
We demonstrate \carving{} on both standard and looped Transformers \citep{dehghani2019universal,fu2026simply}, where a looped Transformer is weight-tied across depth, and applies the same stack of layers in each of $R$ loops. In a looped Transformer, each loop has its own set of gates and learned connections which can also read the outputs from previous loop layers.

Fundamentally, a Transformer layer applies an attention sublayer and then an MLP sublayer, each added to the residual stream $\mathbf z$.
\carving{} scales the output of every sublayer application $u$ (one sublayer in one loop) to the size of the stream and multiplies it by a learned gate $g_u \ge 0$:
\begin{equation}
\mathbf z \leftarrow \mathbf z + g_u\,\frac{\operatorname{rms}(\mathbf z)}{\operatorname{rms}(\mathbf y_u)}\,\mathbf y_u ,
\label{eq:gated-write}
\end{equation}
where $\mathbf y_u$ is the sublayer's output. We initialize each gate at 0.1 (Appendix~\ref{app:gates}). Starting the gates at 0.2 or above removes few or no sublayers and keeps nearly the full inference cost (swept and tested in Appendix~\ref{app:gates}).
This normalization allows the gate alone to set the size of the write relative to the stream, and a gate at zero removes the sublayer's output from the residual stream (Section~\ref{sec:removal}).

The \textit{learned connections} \citep{huang2017densely,pagliardini2024denseformer} allow each layer to reuse and add the outputs of earlier layers to the current residual stream. Each is scaled through a learned coefficient $c$ that gates the connection itself.
Specifically, after layer $k$ with output $\mathbf h_k$ and earlier layer outputs $\mathbf x_j$,
\begin{equation}
\mathbf h_k \leftarrow \mathbf h_k + \operatorname{rms}(\mathbf h_k)\sum_{j} c_{kj}\big(N(\mathbf x_j) - N(\mathbf h_k)\big),
\qquad N(\mathbf x) = \mathbf x / \operatorname{rms}(\mathbf x),
\qquad \|\mathbf c_k\|_1 \le 1 .
\label{eq:connections}
\end{equation}
The coefficients $c_{kj}$ train with the weights and carry no pressure of their own, so each connection grows or shrinks as the loss requires.
The bound $\|\mathbf c_k\|_1 \le 1$ limits the total magnitude of the coefficients from inflating the residual stream through the connections themselves (Appendix~\ref{app:method-details} gives their initialization and how we enforce the bound).

\subsection{The training objective and the value of compute}
\label{sec:price}
\label{sec:goal}

Let $F$ be the training FLOP budget. A sublayer removed at update $t$ makes every later update cheaper, which lowers the FLOPs per token and grants the run more tokens for the same $F$.

Therefore, the critical question is: \textbf{how do we allow \carving{} to balance the quality cost of removing a sublayer with its saved compute?}
Two quantities decide this, which both change as the model learns: what a sublayer contributes to the loss now (which the gradient measures at every update) and what its compute would buy as extra tokens within the training FLOP budget.

To answer this, we combine the two in one training objective and design an adaptive pressure function which eliminates the need for a penalty sweep. At update $t$, the weights $\theta$, the connection coefficients $c$ and the gates $g$ train on the loss
\begin{equation}
\mathcal{L}_t \;=\; L(\theta, c, g;\, B_t) \;+\; \sum_u \pi_u(t)\,\mathbf{1}[g_u > 0],
\label{eq:objective}
\end{equation}
where $L$ is the next-token loss on the batch $B_t$, and every sublayer that stays in the graph carries a pressure $\pi_u$: an \textbf{estimate of the loss those additional tokens would buy if that sublayer was removed}.
This pressure is the penalty on the gates, and serves as the opposing force of the quality loss. Removing $u$ raises the first term by its quality cost and lowers the second by $\pi_u$, so the objective prefers removal exactly when the quality cost is below the pressure.
This moves the compute budget into the loss, replacing the need for a penalty sweep with online estimation instead, which \carving{} does by extrapolating the run's own learning curve \citep{domhan2015speeding}.

Formally, let $L(D)$ be the training loss after $D$ tokens, and let
\begin{equation}
v(D) = -\frac{dL}{d\ln D}
\label{eq:rate}
\end{equation}
be the rate of learning. The loss improves with each multiplication of the token count rather than with each added token, so we measure progress in \textit{e-folds} of tokens: one e-fold multiplies the token count by $e \approx 2.72$, that is, it raises $\ln D$ by one. The rate of learning $v$ is then the reduction in loss per e-fold of tokens (so that doubling the tokens-- $\ln 2 \approx 0.69$ e-folds -- lowers the loss by about $v \ln 2$).
Suppose the run is at $D_t$ tokens and will reach $T$ tokens when the budget is spent with the current model graph. Then, a fraction $\rho_t = (T - D_t)/T$ of its tokens still remains.
If $u$ uses a fraction $\phi_u$ of the training FLOPs per token, the removal of $u$ makes every later token cost $1-\phi_u$ as much, and the run ends at $T\,\big(1 + \rho_t\,\phi_u/(1-\phi_u)\big)$ tokens instead of $T$.
The gained tokens come at the end of the run, where the rate of learning is $v(T)$ per e-fold of tokens, so the pressure on $u$ is $v(T)$ times the e-folds gained:
\begin{equation}
\boxed{\;\pi_u(t) = v(T)\,\ln\!\Big(1 + \rho_t\,\frac{\phi_u}{1-\phi_u}\Big)\;}
\label{eq:price}
\end{equation}
For small $\phi_u$, this reduces to $v(T)\,\rho_t\,\phi_u$: the value of one FLOP at the end of the run, namely $v(T)/(fT)$ with $f$ training FLOPs per token, times the compute $f\,\phi_u (T - D_t)$ that $u$ would use for the rest of the run.
This value of one FLOP is the Lagrange multiplier of the budget in Equation~\eqref{eq:objective}.

By construction, \carving{}'s pressure formulation naturally follows a nice decay schedule (Appendix~\ref{app:learned}).
Early in training, $\rho_t$ is close to one, and compute saved now buys tokens over the whole remaining run.
Late in training, $\rho_t$ falls toward zero (and the pressure with it), because compute saved near the end buys almost no tokens.
Therefore, through $v(T)$, a model that is still learning fast faces a high pressure on its compute, and a model near convergence faces a low one, and it learns to close gates early and less so as training proceeds.
During the final learning-rate decay, we set the pressure to zero.

\subsection{Reading the rate of learning}
\label{sec:reading}

The one quantity the pressure relies on that the run cannot observe directly is the rate of learning at the end of training, $v(T)$.
\carving{} estimates it in two steps: it reads the current rate of learning (the slope of the learning curve) by evaluating the same batch twice, and it forecasts that rate to the end of the run with a power law.

\paragraph{Reading the current rate.}
The slope of the logged training loss is mostly noise (consecutive batches differ in difficulty by more than the loss improves between them). Our trainer therefore keeps a snapshot $\theta_s$ of the parameters from an earlier point $D_s$ and, at a planned update $t$, scores the current batch $B_t$ once more with the snapshot, recording
\begin{equation}
r_t = \frac{L(\theta_s; B_t) - L(\theta_t; B_t)}{\ln(D_t / D_s)} ,
\label{eq:reading}
\end{equation}
where $L(\theta_t; B_t)$ is the loss the training step has already computed on that batch. Both losses see the same tokens, so the difficulty of the batch cancels, and $r_t$ is a finite-difference estimate of the rate $v$ of Equation~\eqref{eq:rate} between $D_s$ and $D_t$.
Each batch calculation costs one forward pass at specific updates (101 to 204 per run), which is a negligible overhead (though also charged to the training FLOP budget). See Appendix~\ref{app:method-details} for the  details and engineering.

\paragraph{Forecasting the rate to the end of the run.}
\label{sec:exponent}
Within a run at a constant learning rate, the loss is well described by a power law in tokens that approaches a floor with exponent $a$ \citep{tissue2024scaling,luo2025multipower}, so the rate is a straight line in log--log coordinates:
\begin{equation}
\ln v(D) = b - a \ln D .
\label{eq:powerlaw}
\end{equation}
We fit this line to the readings by least squares, with a prior on the exponent $a$ taken from published scaling laws \citep{tissue2024scaling,luo2025multipower}, which we replace after approximately $4000$ updates by an exponent fit from the run's own rate of learning (Appendix~\ref{app:method-details}).
The forecast is the constant-learning-rate fit at the projected end of the run, $v(T) = \exp(b - a\ln T)$, and $T$ moves later as sublayers are removed.

\subsection{Applying the pressure to the gates}
\label{sec:gate-update}

\paragraph{The pressure as a penalty on the gates.}
\carving{} trains the gates jointly with the weights. However, the second term of our objective, Equation~\eqref{eq:objective}, is in part a discrete count of the sublayers that remain, so it has no gradient. Therefore, we replace each count $\pi_u\,\mathbf{1}[g_u > 0]$ by the logarithmic penalty $P(g_u) = \frac{\pi_u}{\ln 20}\ln g_u$, which is the smooth replacement for a count of nonzero entries in \cite{candes2008enhancing}  (but unlike \citeauthor{candes2008enhancing}, we add no $\epsilon$ inside the logarithm, so $P$ has no lower bound.)

We provide some intuition for why we apply this construction. Each gate scales how much of the sublayer's write gets written to the stream with a \textit{quality cost}, the first-order increase in loss from removing the sublayer, of $C_u = -g_u\,\partial L/\partial g_u$. This is the loss gradient per \textit{relative} change of the gate; therefore, the right measure of a penalty is how hard it pushes per \textit{relative change of the gate}, $g_u\,P'(g_u)$, which is in the same units of loss as $C_u$. For comparison, a linear ($L_1$) penalty $\lambda g_u$ pushes with $\lambda g_u$ which decreases as the gate shrinks and likewise shrinking the quality cost with it, so the two only compare $\lambda$ with the gradient $\partial L/\partial g_u$ (a coefficient in loss per unit of gate, which would have to be swept). In contrast, the logarithmic penalty pushes with the constant $\pi_u/\ln 20$ at every size, so we compare the pressure with the quality cost directly. Ignoring Adam's momentum, when $\pi_u/\ln 20$ exceeds $C_u$, the gate falls, $C_u$ shrinks with it, and the gate falls all the way until we clip it to exactly zero as in Equation~\eqref{eq:prox}. When $C_u$ exceeds $\pi_u/\ln 20$, the gate is kept and settles at the tension point.

The $\ln 20$ sets how far a gate must fall before the penalty charges its full pressure. In this case, when a gate falls to one twentieth of its value, the penalty decreases by exactly $\pi_u$, because $P(g_u/20) = P(g_u) - \pi_u$. A gate therefore falls only when its quality cost is below $\pi_u/\ln 20$ (about a third of its pressure) which is the rule of Section~\ref{sec:price} applied with a slight margin to keep removal conservative.

We apply our rule with one change to the Adam step. First, the loss gradient takes an ordinary Adam step. Second, we subtract from that the penalty gradient $P'(g_u)$, divided by Adam's denominator $\sqrt{\hat s_u}+\epsilon$ for that gate. If the gate is now below zero, we clip it to zero:
\begin{equation}
g_u \leftarrow \max\!\Big(g_u - \eta\,\frac{\hat m_u}{\sqrt{\hat s_u}+\epsilon} - \eta\,\frac{P'(g_u)}{\sqrt{\hat s_u}+\epsilon},\; 0\Big),
\label{eq:prox}
\end{equation}
with $\hat m_u$ and $\hat s_u$ the bias-corrected first and second moments.
Therefore, the clipping sets a falling gate to exactly zero rather than to a small value that would still cost compute.

\paragraph{Removing sublayers inside the run.}
\label{sec:removal}
Once a gate shrinks to zero, its sublayer contributes nothing, but it still runs until we remove it from the computation graph. Each graph is a static compiled program, so every 300 updates the trainer collects the sublayers whose gates reached zero and compiles the graph without them in a background thread while training continues. 400 updates later, training switches to the new graph. Finally, the  graph at the end of the budget is the final carved model: a static Transformer with a heterogeneous set of layers that runs on any standard kernel.

\section{Experimental setup}
\label{sec:setup}

\textbf{Construction} We apply \carving{} broadly across two types of language models and various scales: standard Transformers at 140M, 370M, 770M, 1.3B and 2.7B parameters, and looped Transformers at 140M, 370M, 770M and 1.3B parameters. In figures and tables, we label each pair by the size of the full-depth model it was carved from.
The standard Transformers follow standard GPT-2/3 construction \citep{radford2019language,brown2020language}, with learned LayerNorm, GELU MLPs, and rotary position encoding \citep{su2021roformer}.
The looped Transformers are Fully Looped Transformers \citep{fu2026simply}, looped four times. Each loop restarts the residual stream from the normalized token embeddings and, from the second loop on, every attention sublayer takes its queries from the previous loop's normalized output to stabilize a known residual-norm explosion with standard looped Transformers.
At each size, the two types share the width, the number of layers and heads (24 layers of width 2{,}048 with 32 heads at 1.3B), rotary position encoding and $4\times$ MLPs. We provide the shapes of every size in Appendix~\ref{app:models}.

\textbf{Baselines} Each \carving{} run compares against two standard model baselines, trained on the same data, the same recipe, and with the same training FLOPs: (1) a ``full-depth'' baseline from which the carved run starts, and (2) a ``shallower-depth'' baseline with fewer layers to match the inference FLOPs per token of the final carved model. We use Muon \citep{jordan2024muon} for the weight matrices and AdamW \citep{loshchilov2019decoupled} for the other parameters, with nanochat's default learning rates (0.02 for standard and 0.0283 for looped Transformers \citep{nanochat}), no warmup, a linear cooldown to zero over the last 20\% of training, and weight decay inversely proportional to the length of the run (all values in Appendix~\ref{app:models}). We train on a shuffled sample of FineWeb-Edu \citep{penedo2024fineweb}, with sequences of 2{,}048 tokens, a 49{,}152-token vocabulary and $2^{20}$ tokens per update. For each scale, we set the FLOP budget $F$ (Section~\ref{sec:price}) to the amount of training FLOPs needed to train each ``full-depth'' baseline for 50 tokens per parameter.

\textbf{Symbolic tasks} We also evaluate \carving{} on a suite of symbolic tasks, which allows us to check whether \carving{} during training removes exactly the computation that the task does not need (known for certain symbolic tasks). To do so, we train a three-layer looped Transformer with four loops, so 24 sublayer applications, on $S_5$ composition (trained length of 8), ListOps, and modular addition modulo 113 \citep{liu2022shortcuts,nangia2018listops,nanda2023progress}. For these tasks, we do not forecast the rate of learning (as they do not have a natural training budget), and instead select a fixed value $\lambda$ which we swept (0.28 for $S_5$, 0.4 for ListOps, and 0.5 for modular addition). Therefore, the pressure on sublayer $u$ is $\pi_u(t) = \lambda\ln\!\big(1 + \rho_t\,\phi_u/(1-\phi_u)\big)$ and the training budget (which sets $T$ and $\rho_t$) is the full-depth model's training FLOPs over about 1.05B tokens. We repeat each run with three seeds. 

\textbf{Pruning and distillation baselines} On the 1.3B standard Transformer, we also compare against pruning the trained full-depth model and retraining it at equal total training compute.
We prune the full-depth model's last save before its cooldown to approximately the same inference FLOPs per token as the carved model and continue training with the remaining budget until it matches the carved run's training FLOPs. We test and compare three types of pruning.
Sheared-LLaMA-style pruning \citep{xia2024sheared} learns hard-concrete masks on whole attention and MLP sublayers under a Lagrangian constraint on the expected inference FLOPs, then continues pretraining.
Minitron-style width pruning \citep{muralidharan2024compact} cuts the hidden and MLP widths uniformly by activation importance on 1{,}024 sequences, keeping every layer and head, then retrains with distillation from the full-depth model plus the language-modeling loss.
ShortGPT-style layer removal \citep{men2024shortgpt} defines a layer's Block Influence (BI) as how much the layer changes its input, and removes the seven layers with the lowest BI. We measure BI on 1{,}024 sequences, and then continue pretraining the rest.

We also train our own distillation baseline at every standard size from 140M to 1.3B and at looped 140M and 370M, by distilling the trained full-depth model of the same size into a freshly initialized student with the shallower-depth shape, whose inference cost matches the carved model's. We train this student on the forward KL divergence to the teacher's next-token distribution, following  \cite{hinton2015distilling}. 

\textbf{Scaling analysis} Following Approach~2 of \citet{hoffmann2022training}, we compare carved and full-depth Transformers of both types at five training budgets from $1.2\times10^{19}$ to $1.2\times10^{20}$ FLOPs. At each budget, we fit a parabola through our trained compute-optimal points to compare compute-optimal sizes and losses, and train the full cooldown of both models at these compute-optimal sizes of each budget. Second, we report two compute multipliers. The \textit{compute multiplier} $k$ of a carved model in Table~\ref{tab:main} is the training compute at which a power law $L(C) = E + A\,C^{-\alpha}$ \citep{henighan2020scaling} fit to the final losses of the trained full-depth models (50 tokens per parameter) reaches the carved loss. The \textit{compute-optimal multiplier} $k^*$ compares their compute-optimal frontiers, that is, the compute at which the full-depth compute-optimal loss reaches the carved compute-optimal loss.

\textbf{Evaluation} We report bits per byte (bpb) on the same 16{,}384 held-out validation sequences for every model, and the CORE score over 22 downstream tasks \citep{li2024dclm,fbt2026fullbandwidth} (Appendix~\ref{app:evaluation}).
On generative tasks across math problems (GSM8K, MATH-500), knowledge recall (TriviaQA), reading comprehension (SQuAD), and last-word prediction (LAMBADA) \citep{cobbe2021training,hendrycks2021measuring,lightman2024lets,joshi2017triviaqa,rajpurkar2016squad,paperno2016lambada}, we sample free-form answers at temperature 0.7 and evaluate pass@$k$ \citep{chen2021evaluating}, the fraction of problems that at least one of $k$ samples solves, with paired 95\% bootstrap intervals over prompts. We compare both the settings of equal sample counts and equal inference budget, where the carved model can draw as many samples as the full-depth model's inference FLOPs allow. We measure decoding throughput on one H100 GPU at contexts of 1{,}024 to 8{,}192 tokens, at equal batch sizes, and at each model's largest tested batch (Appendix~\ref{app:evaluation}). We also compare carved models against publicly released models, which we rescore with the same harness, prompts, few-shot examples, their own tokenizers, and seed as our own models. Appendix~\ref{app:models} has the hardware we used per run.

Finally, we also evaluate whether \carving{}'s lead remains after instruct-tuning our three largest scales (standard 1.3B and 2.7B, looped 1.3B) for one epoch on chat data with nanochat's default recipe \citep{nanochat}. As a check, we repeat the tuning with a gentler recipe, at 0.02 or 0.1 times the pretraining rate with a 3\% warmup, the range used for Llama~2 and T\"ulu~2 \citep{touvron2023llama2,ivison2023camels}, and each model takes the rate with its lower chat validation loss.

\section{Results}
\label{sec:results}

\subsection{\textsc{Carving} produces better models with lower inference cost}
\label{sec:results-pareto}

\begin{table}[!tp]
\centering
\caption{\textbf{\carving{} outperforms standard models in loss and downstream evaluations with up to 34\% fewer FLOPs per token.} Each row compares the full-depth model, a shallower-depth model, the carved model, and a student distilled from the full-depth model into the shallower-depth shape (Section~\ref{sec:results-pruning}). Sublayers per token counts the attention and MLP sublayers a token passes through. All train on the same data and optimizer. The \emph{change} row compares the carved model with the full-depth model, namely the ratios for parameters, sublayers, tokens and FLOPs, differences for bpb, and CORE-22. $k$ measures the compute multiplier, defined as training compute at which full-depth training reaches the carved loss (see Section~\ref{sec:setup}). Finally, Table~\ref{tab:isoflop-minima} gives the compute-optimal multiplier $k^*$.}
\label{tab:main}
\footnotesize
\setlength{\tabcolsep}{3pt}
\renewcommand{\arraystretch}{0.88}
\begin{tabular}{@{}llrrrrrrrrr@{}}
\toprule
& Model & Params & Sublayers & Tokens & Training & Inference & Val. & Equal-token & CORE-22 & $k$ \\
& & (M) & per token & (B) & FLOPs & GFLOPs/token & bpb & bpb & & \\
\midrule
\multicolumn{11}{@{}l}{\emph{Looped Transformer}} \\
140M & full-depth & 151 & 128 / 128 & 7.6 & $3.14 \times 10^{19}$ & 1.18 & 0.899 & 0.983 & 18.9 & \\
 & shallower (15) & 144 & 120 / 128 & 8.0 & $3.14 \times 10^{19}$ & 1.11 & 0.900 &  & 18.9 & \\
 & distilled & 144 & 120 / 128 & 8.0 & $3.14 \times 10^{19}$ & 1.11 & 0.915 &  & 17.9 & \\
 & carved & 151 & 117 / 128 & 7.9 & $3.13 \times 10^{19}$ & \textbf{1.09} & \textbf{0.883} & \textbf{0.958} & \textbf{19.9} & 1.33 \\
\rowcolor{carvetint} & \emph{change} & 1.00$\times$ & \textbf{0.91}$\times$ & 1.04$\times$ & 1.000$\times$ & 0.92$\times$ & $-$0.015 & $-$0.025 & +1.1 & \\
\addlinespace[2pt]
370M & full-depth & 375 & 160 / 160 & 18.8 & $1.92 \times 10^{20}$ & 3.04 & 0.808 & 0.888 & 25.7 & \\
 & shallower (15) & 296 & 120 / 160 & 24.7 & $1.92 \times 10^{20}$ & \textbf{2.31} & 0.816 &  & 24.6 & \\
 & distilled & 296 & 120 / 160 & 24.7 & $1.92 \times 10^{20}$ & 2.31 & 0.828 &  & 23.8 & \\
 & carved & 375 & 124 / 160 & 22.6 & $1.92 \times 10^{20}$ & 2.36 & \textbf{0.788} & \textbf{0.860} & \textbf{28.5} & 1.55 \\
\rowcolor{carvetint} & \emph{change} & 1.00$\times$ & \textbf{0.78}$\times$ & 1.21$\times$ & 1.000$\times$ & 0.78$\times$ & $-$0.020 & $-$0.028 & +2.7 & \\
\addlinespace[2pt]
770M & full-depth & 802 & 160 / 160 & 40.1 & $8.46 \times 10^{20}$ & 6.47 & 0.743 & 0.820 & 31.1 & \\
 & shallower (15) & 622 & 120 / 160 & 53.0 & $8.46 \times 10^{20}$ & 4.89 & 0.750 &  & 29.3 & \\
 & carved & 802 & 118 / 160 & 52.1 & $8.46 \times 10^{20}$ & \textbf{4.76} & \textbf{0.723} & \textbf{0.793} & \textbf{33.5} & 1.62 \\
\rowcolor{carvetint} & \emph{change} & 1.00$\times$ & \textbf{0.74}$\times$ & 1.30$\times$ & 1.000$\times$ & 0.74$\times$ & $-$0.020 & $-$0.027 & +2.3 & \\
\addlinespace[2pt]
1.3B & full-depth & 1,309 & 192 / 192 & 65.4 & $2.25 \times 10^{21}$ & 10.67 & 0.704 & 0.776 & 35.2 & \\
 & shallower (18) & 1,007 & 144 / 192 & 86.7 & $2.25 \times 10^{21}$ & \textbf{8.05} & 0.710 &  & 33.1 & \\
 & carved & 1,309 & 147 / 192 & 82.7 & $2.25 \times 10^{21}$ & 8.17 & \textbf{0.683} & \textbf{0.751} & \textbf{36.0} & 1.71 \\
\rowcolor{carvetint} & \emph{change} & 1.00$\times$ & \textbf{0.77}$\times$ & 1.26$\times$ & 1.000$\times$ & 0.77$\times$ & $-$0.021 & $-$0.025 & +0.8 & \\
\addlinespace[2pt]
\midrule
\multicolumn{11}{@{}l}{\emph{Standard Transformer}} \\
140M & full-depth & 151 & 32 / 32 & 7.6 & $9.12 \times 10^{18}$ & 0.35 & 0.926 & 1.005 & 15.7 & \\
 & shallower (13) & 130 & 26 / 32 & 8.9 & $9.12 \times 10^{18}$ & \textbf{0.30} & 0.931 &  & 16.0 & \\
 & distilled & 130 & 26 / 32 & 8.9 & $9.12 \times 10^{18}$ & 0.30 & 0.947 &  & 14.2 & \\
 & carved & 130 & 27 / 32 & 8.3 & $9.11 \times 10^{18}$ & 0.31 & \textbf{0.922} & \textbf{0.996} & \textbf{16.5} & 1.07 \\
\rowcolor{carvetint} & \emph{change} & \textbf{0.86}$\times$ & \textbf{0.84}$\times$ & 1.09$\times$ & 0.999$\times$ & 0.87$\times$ & $-$0.004 & $-$0.009 & +0.8 & \\
\addlinespace[2pt]
370M & full-depth & 375 & 40 / 40 & 18.8 & $5.28 \times 10^{19}$ & 0.84 & 0.835 & 0.910 & \textbf{22.1} & \\
 & shallower (15) & 296 & 30 / 40 & 24.0 & $5.28 \times 10^{19}$ & \textbf{0.66} & 0.843 &  & 22.0 & \\
 & distilled & 296 & 30 / 40 & 24.0 & $5.28 \times 10^{19}$ & 0.66 & 0.855 &  & 19.8 & \\
 & carved & 296 & 31 / 40 & 22.8 & $5.28 \times 10^{19}$ & 0.67 & \textbf{0.832} & \textbf{0.903} & 22.0 & 1.07 \\
\rowcolor{carvetint} & \emph{change} & \textbf{0.79}$\times$ & \textbf{0.78}$\times$ & 1.22$\times$ & 1.000$\times$ & 0.80$\times$ & $-$0.003 & $-$0.007 & $-$0.0 & \\
\addlinespace[2pt]
770M & full-depth & 802 & 40 / 40 & 40.1 & $2.27 \times 10^{20}$ & 1.74 & 0.769 & 0.842 & 26.7 & \\
 & shallower (15) & 622 & 30 / 40 & 51.9 & $2.27 \times 10^{20}$ & \textbf{1.35} & 0.778 &  & 26.3 & \\
 & distilled & 622 & 30 / 40 & 51.9 & $2.27 \times 10^{20}$ & 1.35 & 0.789 &  & 25.3 & \\
 & carved & 622 & 31 / 40 & 50.1 & $2.27 \times 10^{20}$ & 1.37 & \textbf{0.767} & \textbf{0.836} & \textbf{27.2} & 1.05 \\
\rowcolor{carvetint} & \emph{change} & \textbf{0.78}$\times$ & \textbf{0.78}$\times$ & 1.25$\times$ & 1.000$\times$ & 0.78$\times$ & $-$0.002 & $-$0.006 & +0.5 & \\
\addlinespace[2pt]
1.3B & full-depth & 1,309 & 48 / 48 & 65.4 & $5.93 \times 10^{20}$ & 2.82 & 0.729 & 0.800 & 29.8 & \\
 & shallower (17) & 956 & 34 / 48 & 89.9 & $5.93 \times 10^{20}$ & 2.06 & 0.739 &  & 29.3 & \\
 & distilled & 956 & 34 / 48 & 89.9 & $5.93 \times 10^{20}$ & 2.06 & 0.749 &  & 29.7 & \\
 & carved & 940 & 34 / 48 & 89.3 & $5.93 \times 10^{20}$ & \textbf{2.03} & \textbf{0.727} & \textbf{0.794} & \textbf{31.2} & 1.05 \\
\rowcolor{carvetint} & \emph{change} & \textbf{0.72}$\times$ & \textbf{0.71}$\times$ & 1.37$\times$ & 1.000$\times$ & 0.72$\times$ & $-$0.003 & $-$0.006 & +1.5 & \\
\addlinespace[2pt]
2.7B & full-depth & 2,642 & 64 / 64 & 132.1 & $2.36 \times 10^{21}$ & 5.62 & 0.676 & 0.745 & 36.0 & \\
 & shallower (21) & 1,777 & 42 / 64 & 196.6 & $2.36 \times 10^{21}$ & 3.77 & 0.689 &  & 34.4 & \\
 & carved & 1,751 & 42 / 64 & 196.4 & $2.36 \times 10^{21}$ & \textbf{3.74} & \textbf{0.675} & \textbf{0.741} & \textbf{37.0} & 1.03 \\
\rowcolor{carvetint} & \emph{change} & \textbf{0.66}$\times$ & \textbf{0.66}$\times$ & 1.49$\times$ & 1.000$\times$ & 0.66$\times$ & $-$0.001 & $-$0.004 & +1.0 & \\
\addlinespace[2pt]
\bottomrule
\end{tabular}
\end{table}

\begin{figure}[t]
\centering
\includegraphics[width=\textwidth]{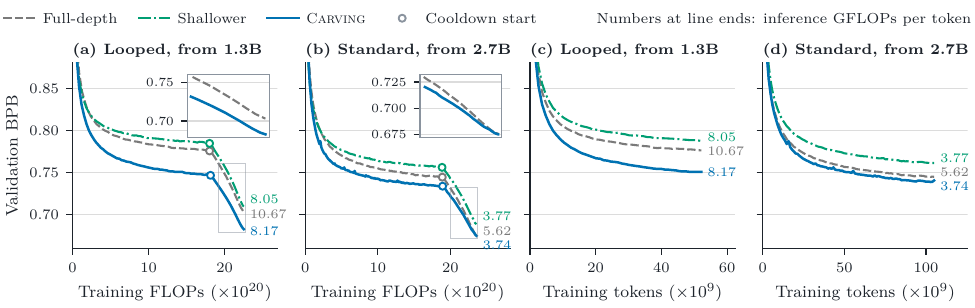}
\caption{\textbf{\carving{} produces models with lower loss than standard models when compared both at isoFLOP and isoToken.}
Validation bpb of the 1.3B looped family and 2.7B standard family against training FLOPs (a, b) and training tokens (c, d).  Panels (c, d) end where the full-depth run starts its cooldown. The title signifies the parameter number of the model \carving{} started from. The number at the end of each line is the model's inference GFLOPs per token.}
\label{fig:training}
\end{figure}

\textbf{\carving{} sets a new compute-quality frontier for both training and inference.}
Across both looped and standard Transformers up to 2.7B parameter scale, \carving{} produces models with lower validation loss than the full-depth model it was carved from at both isoToken and isoFLOP comparisons, while costing 8--34\% fewer FLOPs per token at inference (Table~\ref{tab:main}, Figures~\ref{fig:training} and~\ref{fig:inference}).
Every carved model is also smaller and cheaper to run than the full-depth model it starts from, with 66\% to 86\% of the parameters in the standard Transformer family, and up to 26\% fewer FLOPs per token in the looped Transformer family (Table~\ref{tab:main}).
Moreover, the gains are not simply from the extra tokens per parameter a smaller model gets, as the shallower-depth model ends with higher loss and a lower CORE-22 score than the carved model at all scales and across seeds (Table~\ref{tab:main}). 

\textbf{Carved models perform better on downstream tasks.}
On CORE-22, carved models match or outperform both the full-depth and shallower-depth baselines. The largest gains are noticeably in looped Transformers (Table~\ref{tab:main}; Appendix~\ref{app:downstream}).

On generative tasks, we evaluate both the setting of equal-samples and equal-inference, where given the full-depth model's inference cost, equal-inference means the carved 2.7B standard model can draw approximately 1.5 times as many samples per problem. Therefore, it solves up to 37\% more problems on all five tasks at equal-inference (Table~\ref{tab:passk}). We also report equal samples, within which both models perform near parity at about two points of one another. 

\begin{table}[t]
\centering
\caption{\textbf{At equal inference FLOPs, the carved 2.7B model solves more problems on all five generative tasks.}
Fraction of problems solved by at least one of $k$ samples (pass@$k$) at temperature 0.7, for the standard 2.7B full-depth model versus standard 2.7B carved model. The full-depth model draws 16 samples per problem, whereas the carved model draws 16 (equal samples) or 24, which cost about the same inference FLOPs as the full-depth model's 16 because each carved token costs 0.66 times as much.}
\label{tab:passk}
\footnotesize
\setlength{\tabcolsep}{3pt}
\begin{tabular}{lccccc}
\toprule
 & \multicolumn{3}{c}{Equal samples (16 each)} & \multicolumn{2}{c}{Nearly equal allowance} \\
\cmidrule(lr){2-4}\cmidrule(lr){5-6}
Task & Full-depth & \textsc{Carving} & Difference & \textsc{Carving} (24) & Difference \\
\midrule
GSM8K & 0.284 & 0.300 & $+$0.016 [$-$0.009, 0.040] & \textbf{0.375} & $+$0.091 [0.064, 0.118] \\
MATH-500 & 0.142 & 0.152 & $+$0.010 [$-$0.021, 0.039] & \textbf{0.194} & $+$0.052 [0.018, 0.086] \\
TriviaQA & 0.413 & 0.391 & $-$0.022 [$-$0.029, $-$0.016] & \textbf{0.429} & $+$0.016 [0.009, 0.022] \\
SQuAD & 0.624 & 0.623 & $-$0.001 [$-$0.007, 0.005] & \textbf{0.649} & $+$0.025 [0.019, 0.032] \\
LAMBADA & 0.765 & 0.750 & $-$0.015 [$-$0.024, $-$0.006] & \textbf{0.782} & $+$0.018 [0.008, 0.027] \\
\bottomrule
\end{tabular}
\end{table}

\begin{figure}[t]
\centering
\includegraphics[width=0.8\textwidth]{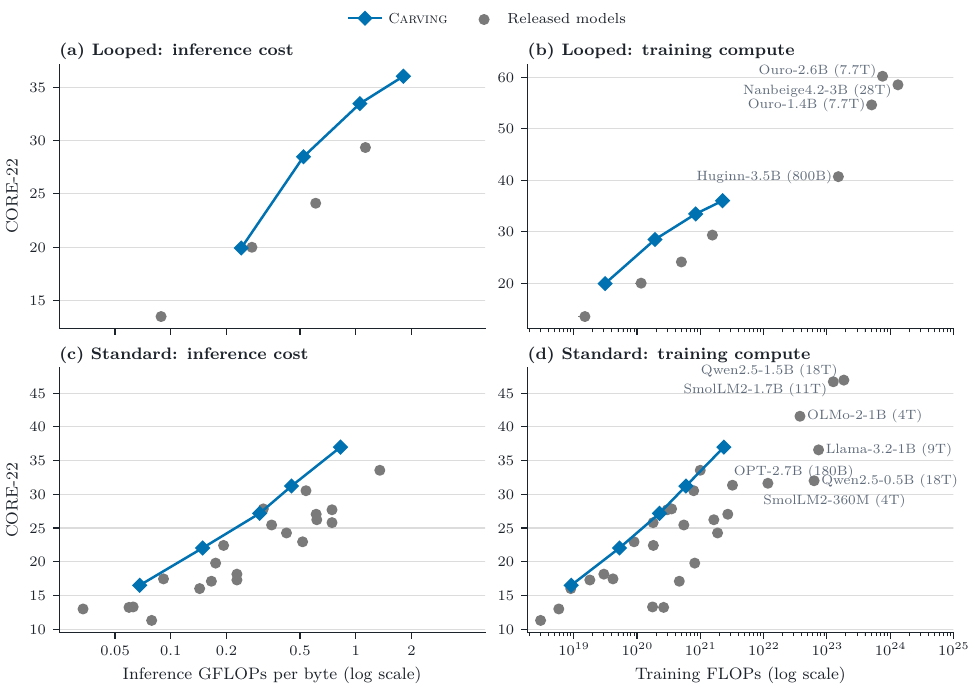}
\caption{\textbf{At matched inference cost, carved models outperform public models trained at similar scales.}
CORE-22 against inference FLOPs per byte (a, c) and training FLOPs (b, d) for looped (a, b) and standard (c, d) Transformers. We label explicitly released models with over 1T training tokens in (b, d). We rescore every released model with our own evaluation where weights are public (Section~\ref{sec:setup}). We list the released models in Appendix~\ref{app:downstream}.}
\label{fig:released}
\end{figure}

\textbf{Carved models beat released models trained with several times their compute.}
We rescore publicly released models with our own evaluation harness, and find that carved models have higher downstream performance than released models trained with significantly more compute (above 800B tokens, otherwise explicitly labeled in Figure~\ref{fig:released}). The standard 1.3B carved model outscores OPT-1.3B and Pythia-1.4B, whose estimated training costs are 2.8 and 4.6 times its own, with fewer inference FLOPs than either, and the 2.7B carved model outscores the larger Delphi-3.4B (Appendix~\ref{app:downstream} contains all comparisons).

\subsection{Carved models decode faster}
\label{sec:results-inference}

\begin{figure}[t]
\centering
\includegraphics[width=\textwidth]{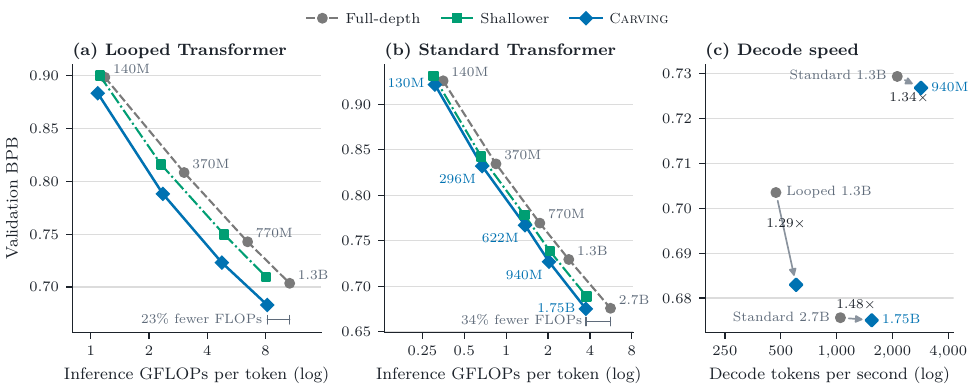}
\caption{\textbf{Carved models have lower loss, cost fewer inference FLOPs, and decode faster.}
Validation bpb against whole-model inference FLOPs per token for (a) looped and (b) standard Transformers. 
(c) Validation bpb against measured decode tokens per second, each model at its largest batch. We decode on one H100 GPU at a 4{,}096-token context, with one compiled call per decode step. Each arrow shows the carved/full-depth throughput ratio (all contexts and batches in Appendix~\ref{app:decoding}).}
\label{fig:inference}
\end{figure}

\textbf{Carved models decode up to 1.45 times faster at equal batch size and up to 1.61 times faster at each model's largest batch.} By construction, \carving{} removes sublayers, and thus runs 8--34\% fewer inference FLOPs per token. 

The removed attention sublayers also shrink the key-value cache, so the carved models keep 5\% to 31\% less KV cache per token, with the largest saving in the 2.7B standard model (Appendix~\ref{app:decoding}). A decode step mostly moves weights and KV cache, and carved models move less of both, which is close to the ratio of their inference FLOPs. On a workstation GPU (RTX PRO 6000, an example edge setting) with one request at a time, the carved standard 1.3B and 2.7B models also decode 1.3 and 1.4 times faster than their full-depth models (Appendix~\ref{app:decoding}). Carved models also gain more at each model's largest batch, because its smaller weights and cache allow for a larger batch (Figure~\ref{fig:inference}c).

\subsection{Carved models outperform pruned models and distilled models}
\label{sec:results-pruning}

We compare \carving{} to the pruning methods of Sheared-LLaMA, Minitron and ShortGPT. We prune the 1.3B standard full-depth model to about the carved model's inference cost using each of these methods, and retrain it to \carving{}'s total training compute (Section~\ref{sec:setup}).
Both Sheared-LLaMA-style pruning with continued training, and Minitron-style pruning (which recovers by distillation from the full-depth model) end with a higher loss and score lower on CORE-22 than the carved model. 
ShortGPT-style layer removal with continued pretraining also ends with a higher loss and scores lower on CORE-22 than the carved model (Appendix~\ref{app:baselines} with the full comparison).

\textbf{At equal training compute, \carving{} outperforms distillation from the full-depth model.}
We distill the trained full-depth model of each pair into a fresh student with the shallower-depth shape (Section~\ref{sec:setup}), so the student has the shallower-depth model's shape and training tokens and the carved model's inference cost. All teacher training and forward pass costs are excluded. At every finished size, the student ends with a higher loss and a lower CORE-22 score than the carved model (Table~\ref{tab:main}) (full numbers in Appendix~\ref{app:baselines}).

\subsection{\carving{}'s advantage grows with scale}
\label{sec:results-scaling}

\begin{figure}[t]
\centering
\includegraphics[width=\textwidth]{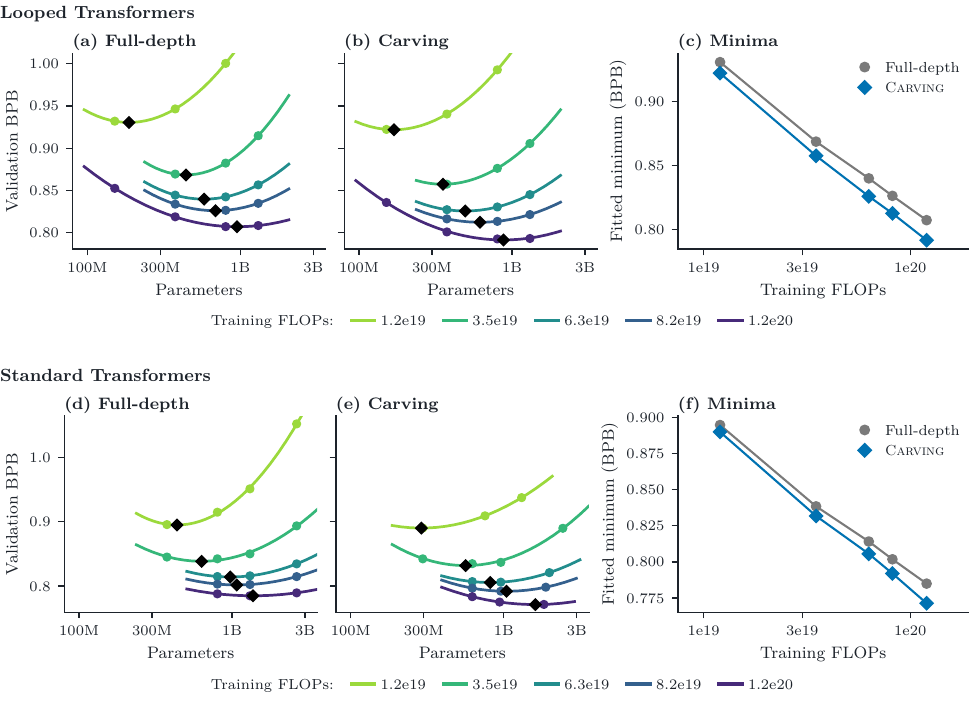}
\caption{\textbf{\carving{}'s advantage grows with scale. The carved compute-optimal loss is below the full-depth compute-optimal loss, and the gap grows with compute.}
Validation bpb post-cooldown for each training budget, from $1.2\times10^{19}$ (light green) to $1.2\times10^{20}$ FLOPs (dark purple), against parameters (final size for carved models), for looped (a-c) and standard (d-f) Transformers. Panels (a, d) show full-depth and (b, e) carved models. We use quadratic fits in log parameters at each budget, and use diamonds to show their minima. (c, f) The minima against training FLOPs.}
\label{fig:isoflop}
\end{figure}

\textbf{\textsc{Carving}'s gains scale with compute (Figure~\ref{fig:isoflop}).}
Following \cite{hoffmann2022training}, we fit scaling laws for \carving{} and calculate the compute-optimal point for carved models versus standard models. At each of our five training budgets, the carved compute-optimal loss is below the compute-optimal loss for non-carved models (Figure~\ref{fig:isoflop}). The gap grows with compute in both families, as the full-depth compute-optimal model needs $k^* = 1.16$ to $1.32$ times the compute to reach the carved compute-optimal loss in looped Transformers, and $1.10$ to $1.25$ times in standard Transformers -- increasing with the budget (Appendix~\ref{app:compute-optimal}).

Moreover, we find that this advantage holds in the regime of overtraining. With about 2.5 times the compute of its main run ($4.9\times10^{20}$ FLOPs, about 130 training tokens per parameter for the full-depth model instead of 50), the 370M looped carved model keeps the same gains over its full-depth baseline (Appendix~\ref{app:compute-optimal}).

In the looped family, the compute multiplier $k$ of Table~\ref{tab:main} also grows with size, from 1.33 at 140M to 1.71 at 1.3B, so \textbf{\carving{}'s gains scale with compute}. At 50 tokens per parameter,  the carved 1.3B looped model matches a full-depth model with about 1.3 times the parameters trained on 1.3 times the tokens. In the standard family, the multiplier is smaller at about 1.07.

\subsection{\carving{} removes computation a task does not need}
\label{sec:symbolic}

\begin{figure}[t]
\centering
\includegraphics[width=0.75\textwidth]{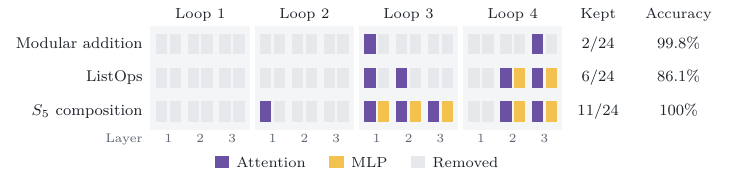}
\caption{\textbf{On symbolic tasks, \carving{} learns a minimal architecture for each task, down to an attention-only Transformer!} Each row is the carved looped Transformer of one task, unrolled into its four loops. The colored sublayers are ones \carving{} kept and gray ones are removed. We label this number as \textit{kept}, as well as the exact match \textit{accuracy}. We show all the seeds in Table~\ref{tab:synthetic}.}
\label{fig:symbolic-arch}
\end{figure}

\textbf{On symbolic tasks, \carving{} finds the minimal circuit that solves the task (Figure~\ref{fig:symbolic-arch}).}
On modular addition, the carved model keeps only 2 of its 24 sublayer applications, both of them attention (an \textit{attention-only Transformer}!). It passes with 99\% test accuracy, and different seeds learned different attention-only sublayers to keep. On ListOps, \carving{} removes 17 to 20 of the 24 sublayers across three seeds. On $S_5$ composition, \carving{} keeps full accuracy at the trained length while removing 13 of the 24 sublayers in all three seeds (Appendix~\ref{app:synthetic}).

\section{Analysis}
\label{sec:analysis}

\subsection{Training without removal}
\label{sec:no-removal}

\textbf{\carving{} matches or beats the performance of DenseFormer-style models with 23\% fewer inference FLOPs.}
\carving{} changes two things relative to the full-depth model. (1) it trains an overparameterized network, with gated sublayer writes and learned connections, and (2) it removes sublayers from the model. For a descriptive comparison, we train the same model without gates and \textit{only learned connections} (as done by \cite{pagliardini2024denseformer}) at standard 140M, 370M and 770M (Figure~\ref{fig:no-removal} in Appendix~\ref{app:learned}). Model Carving achieves about the same performance as the overparameterized DenseFormer control while saving inference FLOPs for free. For instance, the carved looped 370M model beats the DenseFormer-style model in both loss (0.788 against 0.794 bpb) and CORE-22 (28.5 against 27.5) with 23\% fewer inference FLOPs. Therefore, Model Carving achieves the performance of an overparameterized model at the efficiency of a smaller model.

\subsection{The learned topology}
\label{sec:what-removed}
\label{sec:results-structure}

\begin{figure}[t]
\centering
\includegraphics[width=0.9\textwidth]{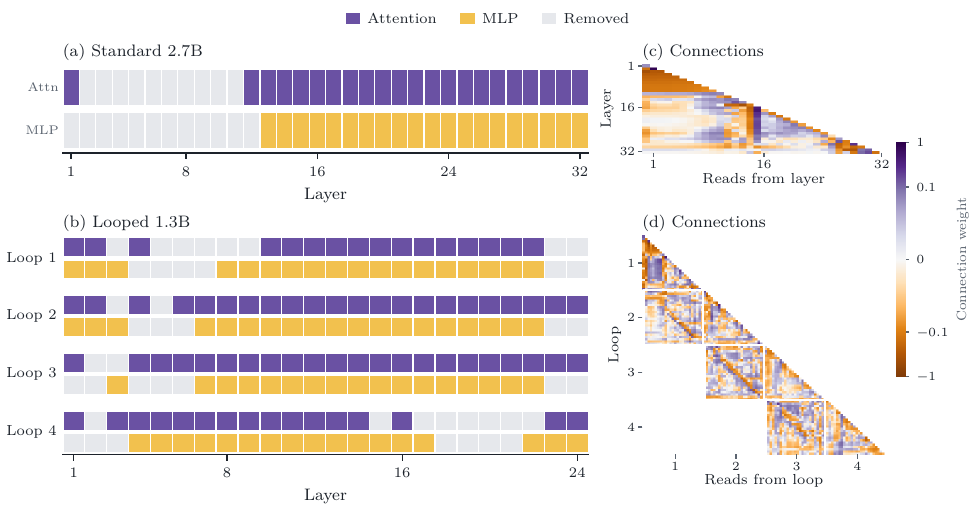}
\caption{\textbf{\textsc{Carving} learns diverse topologies.}
(a, b) The sublayers that the carved standard 2.7B and looped 1.3B models keep (attention, MLP) and remove, by layer and, for the looped model, by loop.
(c, d) Their learned connection weights. Each row is a layer that reads, and each column an earlier output that it can read from. }
\label{fig:topology}
\end{figure}

\textbf{\carving{} learns diverse, yet transferable topologies (Figure~\ref{fig:topology}).}
In every standard Transformer, \carving{} removes sublayers within one early segment. For instance, the 2.7B model removes the attention of layers 2 to 11 and the MLP of layers 1 to 12 (Figure~\ref{fig:topology}a). At every scale, this early segment ends at about a third of the depth (all nine topologies in Appendix~\ref{app:learned}). Interestingly, this is the opposite of what the pruning literature finds, where deeper layers are the easiest to remove \citep{gromov2025ineffectiveness,men2024shortgpt}.

Some useful knowledge would be whether \carving{} removals can also be generally transferable across scales, which would mean we have found a simply more optimal, heterogeneous architecture.
To evaluate this, we predict each sublayer of a held-out size from what \carving{} removed in other sizes. The AUC measures how well these predictions separate removed from kept sublayers (1.0 is perfect and 0.5 is chance), and we find this is at least 0.97 for standard Transformers and looser for looped Transformers (0.76 to 0.88). Therefore, \carving{} seems to have found a better architecture that indeed transfers across scales.

\textbf{\carving{} gives each loop its own depth profile.}
Another benefit of \carving{} is allowing looped models to differentiate their weight-tied iterations without sacrificing the weight-tied property. The 1.3B looped model carves sublayers in the early layers of every loop, at the end of the first three loops, and in the late layers of the last loop (Figure~\ref{fig:topology}b). 
Thus, the same sublayer can be executed in one loop while being removed in another, which gives weight-tied models a different set of sublayers at each depth (which we call the depth-profile).

\textbf{\carving{} retains the sublayers that matter most.}
To test whether \carving{} learns to keep only the ``most important'' sublayers (with respect to predictive quality), we perform a knockout ablation where we remove one sublayer at a time from the final trained model, keep every other weight fixed, and then measure the rise in validation loss.

We find that removing the median kept sublayer of the carved 2.7B model raises the loss \textit{more than five times as much} as removing the median sublayer of the full-depth baseline. In standard Transformers this ratio grows with scale, from about 2.4 times at 140M to 5.5 times at 2.7B.
Therefore, while in the 2.7B full-depth model, most sublayers cost little to lose, as the pruning literature also finds \citep{gromov2025ineffectiveness,men2024shortgpt}, \textbf{almost every sublayer that \carving{} keeps is costly to lose.}

\subsection{The improvement and the learned topology repeat across seeds}
\label{sec:seeds}
We also evaluate a second seed to check whether both the performance of and the topologies learned by \carving{} are robust to seeds. In every pair, both seeds end below the full-depth model. The leads of the first and second seeds are 0.0041 and 0.0038 bpb at standard 140M, 0.0025 and 0.0033 at standard 370M, 0.0026 and 0.0028 at standard 1.3B, and 0.0152 and 0.0156 at looped 140M (Appendix~\ref{app:seeds}).
The sublayer removals and topologies found also repeat across seeds (Appendix~\ref{app:seeds}).

\subsection{\carving{}'s chat lead persists through instruction tuning, and its loss lead through quantization}
\begin{table}[t]
\centering
\caption{\textbf{After instruction tuning, the carved models keep their chat leads under both recipes.} Full-depth and carved models after one epoch of tuning. CORE-22 on the 13 stable tasks (centered points) and chat pass@$k$ (fraction of problems) on GSM8K, TriviaQA and SQuAD, at equal samples (16 each) and within a nearly equal inference allowance (carved models draw samples to match the inference cost of the full-depth model's 16).}
\label{tab:tuned}
\footnotesize
\setlength{\tabcolsep}{3pt}
\begin{tabular}{@{}lrrlrl@{}}
\toprule
 & \multicolumn{3}{c}{Equal samples (16 each)} & \multicolumn{2}{c}{Nearly equal allowance} \\
\cmidrule(lr){2-4}\cmidrule(lr){5-6}
Measure & Full-depth & \carving{} & Difference & \carving{} & Difference \\
\midrule
\multicolumn{6}{@{}l}{\emph{Standard 1.3B, default recipe}} \\
CORE-22 (13 stable) & 33.5 & \textbf{36.3} & $+$2.8 [$+$1.6, $+$3.9] &  &  \\
GSM8K & \textbf{0.459} & 0.444 & $-$0.015 [$-$0.041, $+$0.010] & \textbf{0.496} & $+$0.036 [$+$0.008, $+$0.064] \\
TriviaQA & 0.402 & \textbf{0.409} & $+$0.007 [$+$0.002, $+$0.012] & \textbf{0.435} & $+$0.033 [$+$0.027, $+$0.038] \\
SQuAD & 0.899 & \textbf{0.904} & $+$0.005 [$-$0.000, $+$0.011] & \textbf{0.922} & $+$0.023 [$+$0.018, $+$0.029] \\
\addlinespace[2pt]
\multicolumn{6}{@{}l}{\emph{Standard 1.3B, gentler recipe}} \\
CORE-22 (13 stable) & 40.2 & \textbf{40.9} & $+$0.7 [$-$0.3, $+$1.7] &  &  \\
GSM8K & 0.218 & \textbf{0.232} & $+$0.014 [$-$0.009, $+$0.037] & \textbf{0.277} & $+$0.059 [$+$0.034, $+$0.085] \\
TriviaQA & 0.504 & \textbf{0.506} & $+$0.002 [$-$0.003, $+$0.007] & \textbf{0.530} & $+$0.026 [$+$0.021, $+$0.032] \\
SQuAD & 0.909 & \textbf{0.919} & $+$0.009 [$+$0.004, $+$0.015] & \textbf{0.935} & $+$0.026 [$+$0.020, $+$0.031] \\
\addlinespace[2pt]
\multicolumn{6}{@{}l}{\emph{Standard 2.7B, default recipe}} \\
CORE-22 (13 stable) & 43.5 & \textbf{45.1} & $+$1.6 [$+$0.6, $+$2.6] &  &  \\
GSM8K & 0.525 & \textbf{0.545} & $+$0.019 [$-$0.006, $+$0.045] & \textbf{0.598} & $+$0.073 [$+$0.046, $+$0.101] \\
TriviaQA & 0.504 & \textbf{0.527} & $+$0.023 [$+$0.018, $+$0.028] & \textbf{0.557} & $+$0.053 [$+$0.047, $+$0.058] \\
SQuAD & \textbf{0.936} & 0.931 & $-$0.006 [$-$0.010, $-$0.001] & \textbf{0.946} & $+$0.009 [$+$0.005, $+$0.014] \\
\addlinespace[2pt]
\multicolumn{6}{@{}l}{\emph{Standard 2.7B, gentler recipe}} \\
CORE-22 (13 stable) & \textbf{48.4} & 48.0 & $-$0.5 [$-$1.4, $+$0.5] &  &  \\
GSM8K & 0.318 & \textbf{0.345} & $+$0.027 [$+$0.002, $+$0.051] & \textbf{0.412} & $+$0.095 [$+$0.067, $+$0.123] \\
TriviaQA & 0.603 & \textbf{0.605} & $+$0.002 [$-$0.003, $+$0.006] & \textbf{0.631} & $+$0.028 [$+$0.022, $+$0.033] \\
SQuAD & 0.923 & \textbf{0.925} & $+$0.002 [$-$0.003, $+$0.007] & \textbf{0.942} & $+$0.020 [$+$0.015, $+$0.025] \\
\addlinespace[2pt]
\multicolumn{6}{@{}l}{\emph{Looped 1.3B, default recipe}} \\
CORE-22 (13 stable) & 36.2 & \textbf{42.1} & $+$5.9 [$+$4.6, $+$7.1] &  &  \\
GSM8K & 0.569 & \textbf{0.628} & $+$0.058 [$+$0.032, $+$0.085] & \textbf{0.656} & $+$0.086 [$+$0.059, $+$0.114] \\
TriviaQA & 0.454 & \textbf{0.509} & $+$0.055 [$+$0.050, $+$0.060] & \textbf{0.525} & $+$0.072 [$+$0.066, $+$0.077] \\
SQuAD & 0.932 & \textbf{0.940} & $+$0.007 [$+$0.003, $+$0.012] & \textbf{0.947} & $+$0.014 [$+$0.010, $+$0.019] \\
\addlinespace[2pt]
\multicolumn{6}{@{}l}{\emph{Looped 1.3B, gentler recipe}} \\
CORE-22 (13 stable) & 46.5 & \textbf{50.0} & $+$3.5 [$+$2.4, $+$4.6] &  &  \\
GSM8K & 0.349 & \textbf{0.383} & $+$0.034 [$+$0.009, $+$0.060] & \textbf{0.418} & $+$0.070 [$+$0.042, $+$0.096] \\
TriviaQA & 0.551 & \textbf{0.592} & $+$0.041 [$+$0.036, $+$0.046] & \textbf{0.608} & $+$0.057 [$+$0.051, $+$0.062] \\
SQuAD & 0.930 & \textbf{0.946} & $+$0.016 [$+$0.011, $+$0.020] & \textbf{0.954} & $+$0.024 [$+$0.019, $+$0.029] \\
\bottomrule
\end{tabular}
\end{table}
\textbf{After instruction tuning, the carved models keep their chat leads.}
We tune three model pairs (standard 1.3B and 2.7B, looped 1.3B) for one epoch with nanochat's default recipe (0.8 times the pretraining learning rate) and with a gentler recipe swept over two rates (0.02 and 0.1 times the pretraining rate, with a 3\% warmup), in which each model takes the rate with its lower chat validation loss (Section~\ref{sec:setup}). With the default recipe, carved models outperform non-carved models (CORE-22 over the 13 tasks with stable scores), and on chat pass@$k$ on GSM8K, TriviaQA and SQuAD within the same inference cost. Motivated by the benchmark-noise analysis of \citet{heineman2025signal,madaan2024quantifying}, we call 13 of the CORE tasks ``stable,'' namely those whose scores increase with model size and have a 1.5 signal-to-noise ratio across our full-depth models (Appendix~\ref{app:evaluation}). With the gentler recipe, carved models also beat non-carved models, by 2.0 to 9.5 points of pass@$k$ on GSM8K, TriviaQA and SQuAD (Table~\ref{tab:tuned}). We also try tuning the standard 1.3B pair for two epochs instead of one, as Llama~2 and T\"ulu~2 do \citep{touvron2023llama2,ivison2023camels}, and this widens its CORE-22 lead on the 13 stable tasks from 0.7 to 1.5 points (Appendix~\ref{app:tuning}).

\textbf{Carved models are robust to weight quantization.}
We also evaluate how robust carved models are to quantization. We quantize the weight matrices of every model to 8 bits by rounding, and to 6, 5, 4 and 3 bits with GPTQ \citep{frantar2023gptq} in groups of 64 and calibrated on 128 training sequences, while keeping the embeddings, norms, gates and connection weights in full precision. \carving{}'s lead is robust to weight quantization down to 4 bits (except standard 2.7B), and down to 3 bits in the looped family (Appendix~\ref{app:quantization}).

\subsection{Cooldown vs.\ excess loss}
\label{sec:cooldown}

\textbf{Carved models have lower loss excess from the constant-LR phase.}
One observed phenomenon is that carved models gain less from their cooldown than full-depth models. Notably, across the eight pairs from 140M to 1.3B, \carving{}'s final cooldown lowers the loss 10\% to 16\% less than the full-depth model's, while the shallower-depth model's cooldown gains as much as the full-depth model's, ruling out the size confound (Appendix~\ref{app:cooldown}).

A cooldown fundamentally removes the excess loss that the noise of a constant learning rate adds \citep{hagele2024scaling,wen2025understanding}, and  we find that carved models carry less of that excess. We test this by averaging the weights along the constant-rate trajectory to remove much of this noise. At standard 140M, we find this averaging helps \carving{} less, by 4\% to 9\% (Appendix~\ref{app:cooldown}).
Therefore, part of \carving{}'s lead at the peak learning rate is gain that the full-depth model collects only during its cooldown.

To test whether the shape or length of the cooldown has an effect, we rerun the final cooldown of the standard 140M pair with cosine and square-root shapes, and with a longer cooldown (about 35\% and 45\% of training instead of 20\%) in both runs. Under every schedule, \carving{}'s cooldown gain stays 10 to 13\% smaller than the full-depth model's, while still finishing better overall.

\section{Related work}
\label{sec:related}

\paragraph{Learning the architecture.}
The central idea of \textsc{Model~Carving} is to learn which sublayers a Transformer should keep in the same run that trains its weights. We take inspiration from prior work with a similar objective but using some form of a sparsity penalty, as numerous approaches indeed apply $L_1$, $L_0$, and FLOP-based penalties on channels and heads \citep{liu2017learning,gordon2018morphnet, louizos2018learning}.   Structure can also be learned during training without a penalty, as \cite{zhu2017prune} used a gradually increasing pruning schedule to sparsify the weights during training. Although these methods are adjacent, the main difference is both \carving{}'s penalty and dynamic sublayer removal.  

Neural architecture search likewise pursues learning a more capable architecture, but employs a second optimization problem. Early methods trained a space of thousands of candidate networks and performed search \citep{zoph2017neural,real2019regularized,so2019evolved}, while DARTS, ProxylessNAS, and FBNet optimize architecture parameters on held-out data alongside a supernet's weights followed by a fresh retraining \citep{liu2019darts,cai2019proxylessnas,wu2019fbnet}. \textsc{Model~Carving} needs neither the second optimization search nor the retraining.

\paragraph{Gates and connections across depth.}
\carving{} builds on work that scales and gates how layers write to the residual stream. Within the realm of skip connections, Highway networks gate each skip connection with learned, input-dependent gates \citep{srivastava2015training}, residual networks use identity skip connections \citep{he2016deep}, and ReZero scales each residual branch by a learned scalar that starts at zero \citep{bachlechner2020rezero}. Within the realm of over-connectivity, DenseNet connects each layer to all earlier ones \citep{huang2017densely}, and DenseFormer and hyper-connections learn how much each layer reads from earlier layers \citep{pagliardini2024denseformer,zhu2024hyperconnections}. \textsc{Model~Carving} uses both ideas, with a scalar gate on each sublayer and DenseFormer-style learned connections. 

\paragraph{Pruning and distillation.}
In 1989, \textit{Optimal Brain Damage} \citep{lecun1989optimal} showed how unimportant weights from a trained network could be removed with little impairment. \carving{}'s thesis pushes further, namely that removal of unimportant sublayers allows critical sublayers to train better. However, most methods for pruning derive the mask from a larger model that has already been trained. For structured pruning, much of the benefit comes from the pruned architecture rather than from the inherited weights \citep{liu2019rethinking}, which suggests finding that architecture during training instead. In language models, Sheared-LLaMA and CoFi learn pruning masks under $L_0$ regularization and Lagrange multipliers \citep{wang2020structured} that prune toward either a target shape or desired target sparsity \citep{xia2024sheared, xia2022structured}, Minitron prunes by an estimated importance heuristic and retrains with distillation \citep{muralidharan2024compact}, and it is well known that whole layers can be removed with little loss \citep{men2024shortgpt,gromov2025ineffectiveness,he2024attention}.  \textsc{Model~Carving} removes sublayers without a specified target, and only in tension with the prediction loss (Appendix~\ref{app:learned}). Carved models also remove sublayers in roughly the first third of each standard Transformer, in contrast to layer-removal studies that have found deeper layers easier to remove.

\paragraph{Adaptive computation.}
Another way to save computation is to decide per token, at inference-time, how much of the network to run, through halting \citep{graves2016adaptive}, mixtures of experts \citep{shazeer2017outrageously}, early exits \citep{schuster2022confident}, or routers that skip blocks \citep{raposo2024mod} or set how many times a shared stack runs \citep{bae2025mor}. However, these methods keep each parameter and run a router or halting rule while decoding, whereas a carved model is a single static graph and stores no key--value cache for removed attention sublayers. In fact, the two are complementary!

\paragraph{Scaling laws.}
Lastly, compute-optimal training picks one model size for a budget before training, from loss fits across many runs \citep{kaplan2020scaling,hoffmann2022training}, while accounting for inference favors smaller models trained on more tokens \citep{sardana2024beyond}. \textsc{Model~Carving} makes this trade inside one run and brings it to each sublayer by extrapolating the run's own learning curve \citep{domhan2015speeding} with the power-law form of within-run loss laws \citep{tissue2024scaling,luo2025multipower}.

More broadly, the methods above set the trade-off between quality and compute outside the run that trains the model, through a penalty coefficient, a target size, a routing capacity, or a scaling-law fit. The thesis of \textsc{Model~Carving} is to place this in compute versus prediction quality tension within the training itself, which is why one set of fixed settings works at every size and in both model families.

\section{Discussion}
\label{sec:discussion}

\textbf{\textsc{Carving} keeps its lead while removing computation early in training, contrary to pruning.}
Pruning at initialization or early in training is far from a new idea. These masks learned before training, such as in SNIP \citep{lee2019snip}, GraSP \citep{wang2020picking} and SynFlow \citep{tanaka2020pruning}, often fall short of pruning after training \citep{frankle2021pruning}. However, despite \carving{} making most of its sublayer removals during the first 20\% of training, carved models reach a lower loss than the full-depth model at matched training compute.
We discuss below why \carving{} differs in this case. 

First, \textsc{Carving} removes full sublayers, akin to structured pruning, for which most of the gain comes from the pruned architecture rather than the individual weights \citep{liu2019rethinking}. \carving{} thus finds this architecture inside the run that trains it.

Second, the gates update smoothly while the weights learn. As in ReZero \citep{bachlechner2020rezero}, each sublayer writes through a scalar gate. Our pressure pushes every gate toward zero, but a gate falls only where the pressure outweighs its sublayer's contribution, while the rest of the network keeps training. Therefore, a gate can fall far and rise again, yet only a zero gate removes its sublayer, and no sublayer with a gate value of zero returns to a nonzero gate value (Appendix~\ref{app:gates}).

Finally, we give the gates their own optimizer, with the pressure as their only sparsity penalty. The gates take a separate Adam step without weight decay, and a normalized write lets the gate alone set the size of each write relative to the stream (Equation~\ref{eq:gated-write}). Therefore, a gate only adjusts by the loss and the pressure.

\textbf{\carving{} offers a compute-efficient solution for building edge models.}
Edge devices are deployed in settings where inference compute and memory are the bottlenecks. Currently, the most common ways to build edge models involve pruning a larger model or distilling from it. \carving{} offers a new method to develop and train smaller models directly, also robust to quantization.  

\section{Limitations and future work}

\paragraph{Larger scale.}
In this study, we explore the largest scales of 2.7B parameters on standard Transformers and 1.3B parameters on looped Transformers. This regime demonstrates that the \textit{percentage} of inference compute that Model Carving saves increases with scale (see Section~\ref{sec:results-scaling}). Deeper models may have even more potential to be carved, and thus an immediate follow-up is to test our method on larger scales, longer training, and different data.

\paragraph{Cooldown.}
The pruning pressure forecasts the learning with the assumption of a fixed LR, but we simply turn off the pressure during the cooldown (with intuition being to settle into a minimum). As a result, Model Carving gains 10\% to 16\% less from the cooldown than dense models do (see Section~\ref{sec:cooldown}), given that the pruning pressure does not account for the cooldown yet. We suspect it would be more accurate to use a pruning pressure that takes into account the learning-rate schedule \citep{tissue2024scaling, luo2025multipower}, and that in future work, it is also worth exploring whether pruning the structure during the cooldown is fruitful.

\paragraph{Initial settings.}
Most of the pruning happens in the first 20\% of the training, which indeed relies on the prior. The gate initialization and margin $\ln 20$ are hyperparameters applied at all scales and families. In future work, it would be interesting to explore uncertainty-aware learning predictors (to push pruning until the estimate is stable), dynamic pruning rules (that allow the re-insertion of pruned sublayers), and a theoretical derivation of these hyperparameters.

\paragraph{Dynamic batch sizes and learning rate.}
In a similar vein, we believe that \carving{} may benefit from \textit{dynamic hyperparameters}, as carving out sublayers alters model size and thus the optimal hyperparameters. Scaling laws (with respect to model size and training tokens) are well-defined for adapting learning rate, batch size, and weight decay. For instance, under the maximal update parametrization ($\mu$P), the optimal learning rate remains stable as the model's width changes \citep{yang2021tensor}, or optimal batch size increases with training tokens \citep{bergsma2025power}. We designed a first schedule that increases the batch size during training \citep{smith2018dont,mccandlish2018empirical}, which reached the same final loss of the carved 140M model with 38\% fewer updates, and we plan to test it at larger sizes.

\paragraph{Granularity and deployment costs.}
Model Carving currently prunes entire attention and MLP applications. More generally, however, our method can prune any granular computational component that has a gating mechanism and a fraction of the FLOPs. It would be interesting to explore the use of Model Carving on attention heads and MLP dimensions, layer types, and to include device latency or memory in the pruning pressure.

\paragraph{Combining with adaptive computation.}
Model Carving arrives at a static structure and is not able to adaptively allocate more compute to harder tokens. Token-level adaptive computation at inference time (e.g., through halting, early exits, and routing \citep{graves2016adaptive, schuster2022confident, raposo2024mod, bae2025mor}) is a complementary approach to Model Carving. Specifically, Model Carving first arrives at a structure, and then adaptive computation could be used to allocate token-level compute based on the sublayers (or number of loops in a looped Transformer).

\paragraph{Carving pretrained models.}
All our experiments train from scratch and from a random initialization. In future work, it would be interesting to explore whether principles from Model Carving can be used to prune or fine-tune a pretrained model by adding gating and training with the pruning pressure.

\bibliography{references}
\bibliographystyle{iclr2027_conference}

\appendix
\section{Models and training settings}
\label{app:models}

We list the shape of every model in Table~\ref{tab:shapes}. At each scale,  standard and looped Transformers share the same width, number of layers and heads, a 49{,}152-token vocabulary, a 2{,}048-token context, MLPs four times the width with GELU, learned LayerNorm, and rotary position encoding (the only difference being a looped Transformer applies its layers four times).

\begin{table}[h]
\centering
\caption{Model shapes and the full-depth training runs. Tokens and updates are measured with respect to the full-depth runs, at 512 sequences of 2{,}048 tokens per update.}
\label{tab:shapes}
\footnotesize
\begin{tabular}{lrrrrrr}
\toprule
Size & Layers & Width & Heads & Parameters (M) & Training tokens (B) & Updates \\
\midrule
140M & 16 & 768 & 12 & 151 & 7.6 & 7{,}201 \\
370M & 20 & 1{,}152 & 18 & 375 & 18.8 & 17{,}889 \\
770M & 20 & 1{,}728 & 27 & 802 & 40.1 & 38{,}223 \\
1.3B & 24 & 2{,}048 & 32 & 1{,}309 & 65.4 & 62{,}401 \\
2.7B & 32 & 2{,}560 & 32 & 2{,}642 & 132.1 & 126{,}001 \\
\bottomrule
\end{tabular}
\end{table}

\textbf{Optimizer settings} We use Muon \citep{jordan2024muon} for the weight matrices with momentum 0.95 and five Newton--Schulz steps, and AdamW \citep{loshchilov2019decoupled} for the embeddings, unembedding, and norms with $\beta_1 = 0.8$ and $\beta_2 = 0.95$. The learning rate is 0.02 for standard and 0.0283 for looped Transformers, with the embedding at 0.15 times that rate and the unembedding at 0.2 times that rate, no warmup, and a linear cooldown to zero over the last 20\% of the updates. We set weight decay to be inversely proportional to the length of the run (calculated from the learning rate times the weight decay equals 12.5 divided by the planned number of updates) which gives 0.087 at standard 140M and 0.010 at standard 1.3B. Carved runs adjust this value  as sublayers leave (Appendix~\ref{app:method-details}).

\textbf{Hardware} The runs trained on NVIDIA H100, H200 and B200 GPUs of a university cluster, a cloud provider, and an industry research cluster.  Every run used 1{,}048{,}576 tokens per update.

\section{Method details}
\label{app:method-details}

\paragraph{One training update.}
(1) Run the forward and backward pass on the batch with the current graph and update the weights and connections. (2) Compute the rate of learning $r_t$ of Equation~\eqref{eq:reading} from the snapshot $\theta_s$ and add it to the fit of Equation~\eqref{eq:powerlaw}. (3) Project the end of the run $T$ from the remaining budget and the cost per update of the current graph, and set the end rate $v(T) = \exp(b - a \ln T)$. (4) For each open gate, use Equation~\eqref{eq:price} to set its pressure, take the Adam step, and then the penalty step and the clip at zero of  Equation~\eqref{eq:prox}. (5) At each removal step, remove the sublayers whose gates are at zero in the executed graph. (6) Charge the executed FLOPs and stop when the charge reaches the budget.

\paragraph{Cost model.}
To calculate FLOPs per token, we count every executed sublayer application, connection, and readout, and training FLOPs are three times the forward FLOPs as is conventional. An attention application costs $24d^2 + 12Td$ forward-and-backward FLOPs per token at width $d$ and context $T$, and an MLP application $48d^2$. $\phi_u$ in Equation~\eqref{eq:price} is sublayer $u$'s share of the training FLOPs per token of the executed graph without the sublayers whose gates are already at zero.

\paragraph{Projected end.}
At each update, we project the last update from the remaining budget and the cost per update of the current model graph. Here, $T$ in Equation~\eqref{eq:price} is the token count at that update, and $\rho_t$ is the fraction of $T$ that is still remaining.

\paragraph{Snapshots.}
We use two snapshots to retain the weights and gates in float32 for updating the learning exponent. We take a snapshot from an earlier point $D_s$ which is a quarter to half an e-fold of tokens back.

\paragraph{Exponent fit.}
The fit of Equation~\eqref{eq:powerlaw} places each snapshot reading (forward pass value) at the midpoint $x_t = \tfrac12(\ln D_s + \ln D_t)$ of its interval and solves, by least squares with a prior $a_0$ on the exponent from published loss curves,
\begin{equation}
(a, b) = \arg\min_{a \ge 0,\, b}\; \sum_t \big(\ln r_t - b + a x_t\big)^2 + \Lambda (a - a_0)^2 .
\label{eq:fit}
\end{equation}
The prior's weight $\Lambda$ in Equation~\eqref{eq:fit} equals the data's weight on the slope at the first fit, so the readings outweigh it as they accumulate. We remove readings from an early transient period, and the fitting starts around update 1{,}400 at 1.3B. The exponents the runs learn end between 0.4 and 0.65, which is within or very closely near the 0.42 to 0.62 reported for single-run loss curves by \cite{tissue2024scaling,luo2025multipower} (Appendix~\ref{app:learned}).

\paragraph{Gates.}
Every gate starts at 0.1 and takes Adam steps with learning rate $10^{-3}$ per $2^{20}$ tokens, $\beta_1 = 0.9$ and $\beta_2 = 0.99$. We give a looped model one gate per sublayer per loop, so a sublayer decreases the number of stored parameters only when it is closed in each loop.

\paragraph{Removal schedule.}
Every 300 updates, the trainer compiles a smaller execution graph in a background thread without the sublayers whose gates are zero, and switches to it 400 updates later.

\paragraph{Connections.}
Each connection row starts with a sum of 0.5 and trains with Adam at learning rate $10^{-3}$, $\beta_1 = 0.9$ and $\beta_2 = 0.99$. After each update, a row with $L_1$ norm above one is rescaled to norm one, and a row with signed sum above 0.95 is rescaled to 0.95. The connections themselves carry no pressure.

\paragraph{Weight decay and cooldown.}
We normalize the weight decay to the projected length of the run (Appendix~\ref{app:models}) which adjusts as sublayers leave. The cooldown covers the last 20\% of the projected run, after which we set the pressure to zero and no sublayers are removed.

\paragraph{Fixed settings.}
Apart from the pressure, \carving{} has a short list of fixed ``settings'' for both families and all scales: the initialized gate value (0.1), the update at which the pressure starts (340), the penalty constant $\ln 20$, the removal intervals, and the cooldown (20\%).

\paragraph{The carved model at inference.}
After training, the carved model is a static Transformer with fewer (and heterogeneous) sublayers. The same graph runs for every token, so serving does not require any routing or masking, and runs on any standard kernel. Further, each kept attention application keeps the standard key--value cache. The learned connections only act as gates to the current token's earlier layer outputs, so they add no state across tokens.

\subsection{Gate falling and rising behavior}
\label{app:gates}

\begin{figure}[h]
\centering
\includegraphics[width=\textwidth]{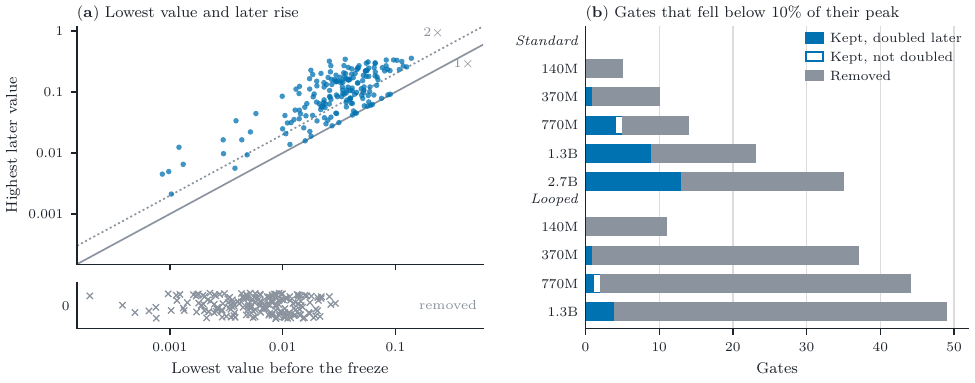}
\caption{\textbf{Gates falling and rising behavior, and a gate at zero stays at zero.} (a) For every gate of the nine carved runs, its lowest reading before the cooldown freeze against its highest later reading. Crosses: the removed sublayers whose gates reach zero and do not return. (b) The gates that fell below a tenth of their peak, by run.}
\label{fig:gate-reopening}
\end{figure}

Notably, once a sublayer is removed (gate shrinks to zero), it does not return. Across the nine carved runs, 193 sublayers were removed before the cooldown freeze and none returned. Gates above zero do fall and rise (Figure~\ref{fig:gate-reopening}), as of the 671 kept gates, 184 fell below half of their peak and 141 of them later doubled, and 35 fell below a tenth of their peak and 33 of those doubled. Once a gate falls nearly to zero, however, zero follows quickly by our pressure construction.

\paragraph{Starting gate value.}
Table~\ref{tab:gate-start} shows a sweep over the value at which every gate starts for the 140M model in both families. Initializing the gate at 0.2 and above ends up removing few or no sublayers while keeping nearly the full inference cost. Our paper default 0.1 removes 3 to 5 of 32 applications in the standard runs and 11 of 128 in the looped run.

\begin{table}[h]
\centering
\caption{Starting gate value at 140M. Removed: sublayer applications removed by the end of training, of 32 in the standard model and 128 in the looped model. FLOPs per token: the carved model's inference FLOPs as a fraction of the full-depth model's. Lead: the carved model's lead in bpb over the full-depth model.}
\label{tab:gate-start}
\footnotesize
\begin{tabular}{lrrrrrr}
\toprule
 & \multicolumn{3}{c}{Standard 140M} & \multicolumn{3}{c}{Looped 140M} \\
\cmidrule(lr){2-4}\cmidrule(lr){5-7}
Starting gate value & Removed & FLOPs per token & Lead & Removed & FLOPs per token & Lead \\
\midrule
0.05 & 3 & 0.90 & 0.0016 to 0.0034 & 21 & 0.83 & 0.0148 \\
0.1 & 3 to 5 & 0.84 to 0.90 & 0.0005 to 0.0043 & 11 & 0.91 & 0.0149 \\
0.2 & 0 to 2 & 0.93 to 1.00 & 0.0070 to 0.0076 & 2 & 0.99 & 0.0161 \\
0.4 & 0 & 1.00 & 0.0100 to 0.0110 & 3 & 0.98 & 0.0178 \\
\bottomrule
\end{tabular}

\end{table}

\section{What the runs learned}
\label{app:learned}

\begin{figure}[h]
\centering
\includegraphics[width=\textwidth]{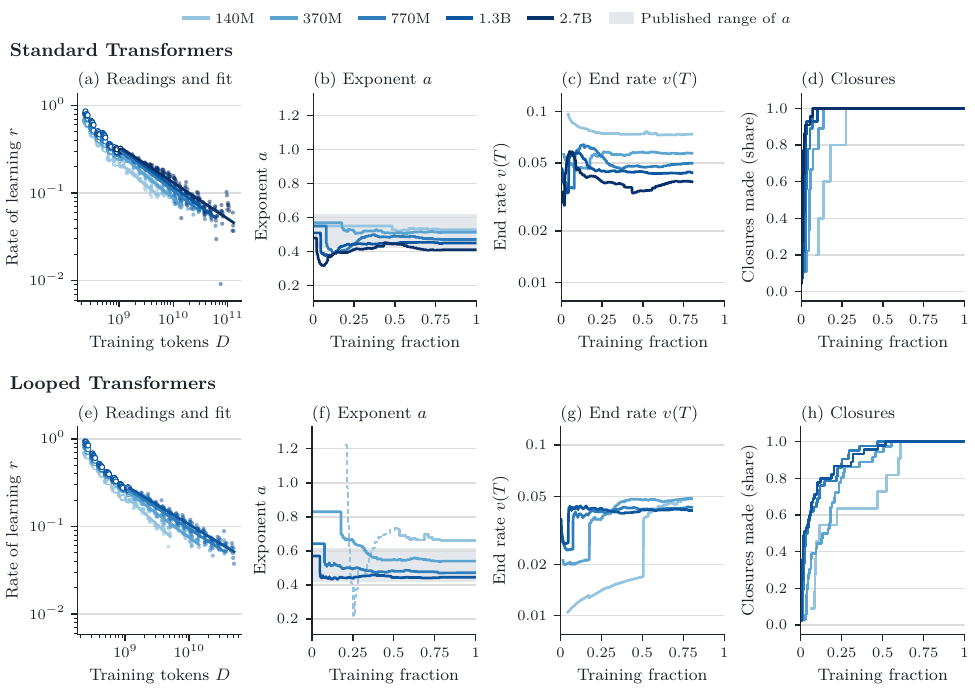}
\caption{\textbf{How the pressure follows the learning curve in \carving{}.} Rows: standard (top) and looped (bottom) Transformers, and color representing model size. (a, e) Paired readings of the rate of learning against training tokens, and each run's own fit at the end of training. (b, f) The exponent over training with the gray band indicating the published exponent values. (c, g) The gate pressure set by the forecasted rate of learning at the end of the run, $v(T)$ (Equation~\eqref{eq:price}). The pressure continues until the cooldown freeze at 80\% of training. (d, h) The share of the final removals made by each training fraction.}
\label{fig:carving-process}
\end{figure}

\begin{figure}[h]
\centering
\includegraphics[width=\textwidth]{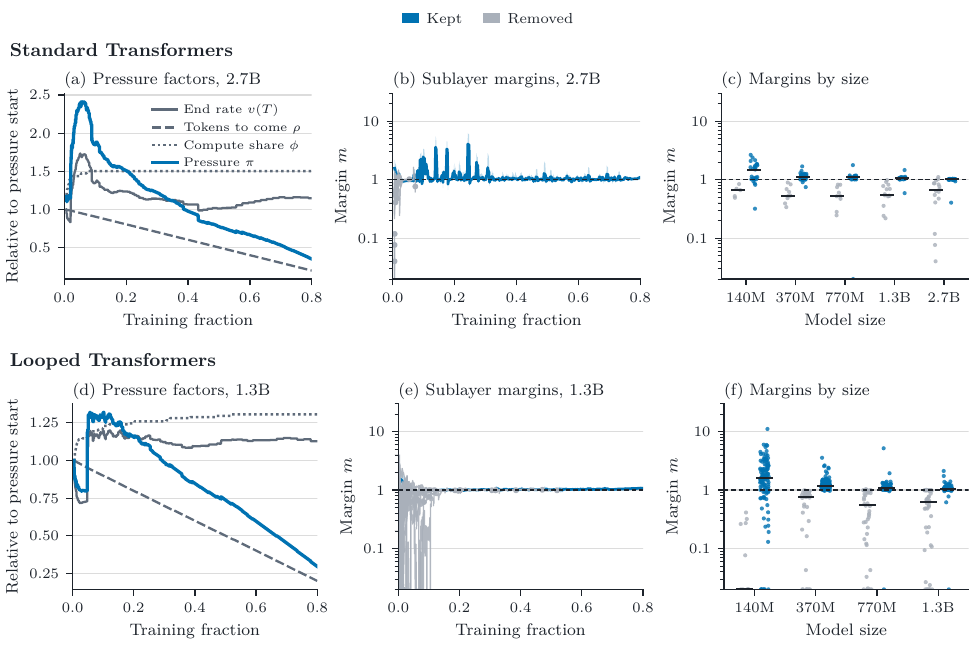}
\caption{\textbf{The pressure on each sublayer, and the margin for removal.} (a, d) The factors of the pressure for the largest model of each family, namely the end rate $v(T)$, the tokens to come $\rho$, the compute share $\phi$ of an MLP sublayer, and its pressure $\pi$. (b, e) The margin over training, from means over the last 1{,}000 updates. (c, f) For each size, the margin of each removed sublayer just before its removal (gray) and the median margin of each kept sublayer over 60 to 80\% of training (blue), and the bars being the medians.}
\label{fig:carving-pressure}
\end{figure}

We show the ``learning'' process of \carving{} in Figure~\ref{fig:carving-process}. The readings of the rate of learning lie on one power law per run, and indeed the exponent each run learns settles within or very closely near the published range. As sublayers leave, the projected end of the run moves later by design and the trainer rescales the pressure on every sublayer, and, without a schedule set by hand, the pressure falls to under a third of its starting value before the cooldown. The removals follow the rule of Section~\ref{sec:gate-update}, namely a sublayer's margin is its estimated contribution to the loss over its pressure (Figure~\ref{fig:carving-pressure}).

\begin{figure}[h]
\centering
\includegraphics[width=\textwidth]{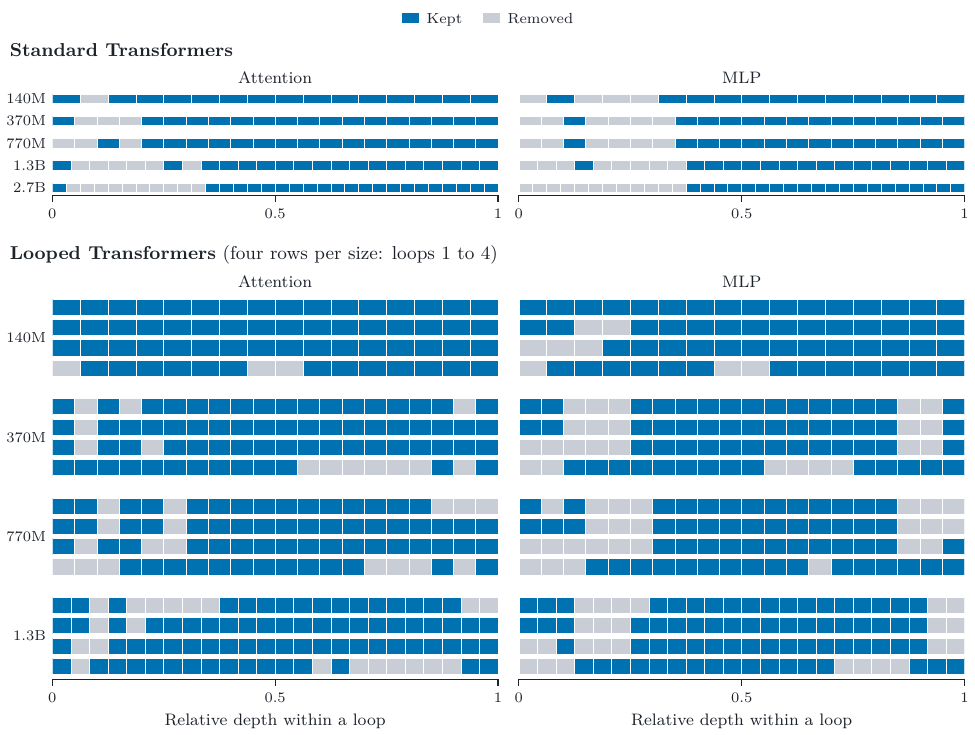}
\caption{\textbf{The learned topology of all nine carved models.} Blue: kept sublayer; gray: removed. The horizontal axis is relative depth within a loop, so models of different depth align. A standard model is one row, and a looped model is four rows with its loops 1 to 4 from top to bottom.}
\label{fig:topology-maps}
\end{figure}

\begin{figure}[h]
\centering
\includegraphics[width=\textwidth]{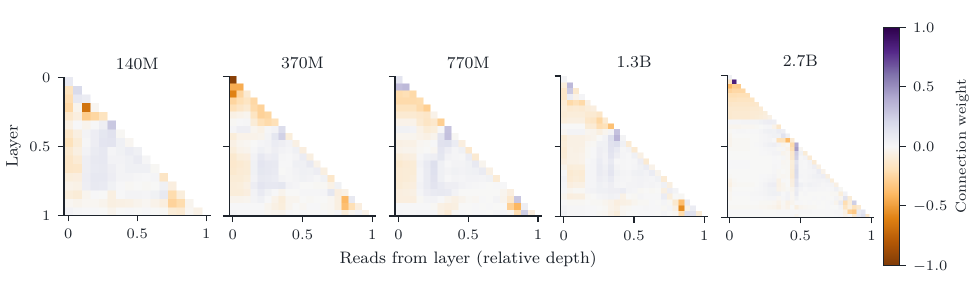}
\caption{\textbf{Learned connection weights of the standard carved models.} Reading layer goes down, source layer goes across (0: the embedding), and both on relative depth. Purple: the layer input moves toward that source; orange: away from it. The looped models also read the outputs of the previous loop, which we did not draw.}
\label{fig:connections}
\end{figure}

Figure~\ref{fig:topology-maps} shows the final topology of every carved model and Figure~\ref{fig:connections} the learned connection weights of the standard models. Interestingly, in every standard model, the largest connection weights sit in the first layers' reads of the embedding.

\begin{figure}[t]
\centering
\includegraphics[width=\textwidth]{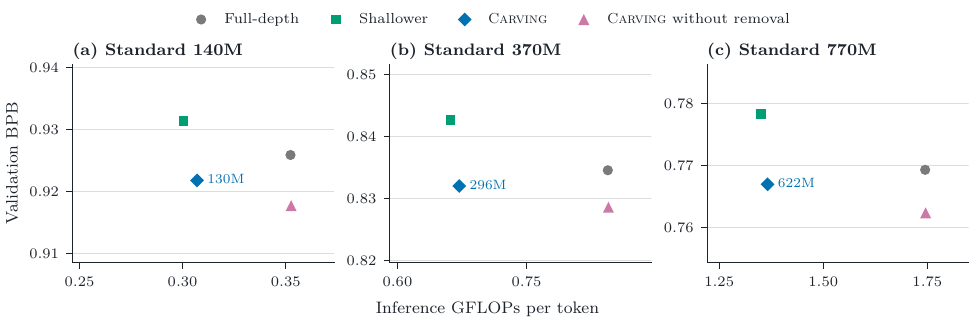}
\caption{\textbf{Carved models run up to 22\% fewer inference FLOPs than the DenseFormer-style models.} Validation bpb against whole-model inference FLOPs per token comparing \carving{} vs. \carving{} without removal (DenseFormer controls). Labels: the carved model's parameter count.}
\label{fig:no-removal}
\end{figure}

\paragraph{The same network without removal.}
Figure~\ref{fig:no-removal} and Table \ref{tab:no-removal} compare each standard carved model with the same overparameterized model with learned connections but without gate removals, trained for the same FLOPs. These DenseFormer-style models have the same inference cost as the full-depth model's plus the connections, which is under 0.1\% more FLOPs per token. They end about 0.006 bpb below full-depth models and 0.014 bpb below shallower-depth models. On CORE-22, the DenseFormer model scores 22.0 and 27.4 at 370M and 770M, against 22.0 and 27.2 for the carved models and 22.1 and 26.7 for the full-depth models. Therefore, as mentioned in Section \ref{sec:analysis}, Model Carving achieves about the same performance as the overparameterized DenseFormer control while saving inference FLOPs for free -- 22\% (standard) and 23\% (looped) less inference FLOPs comparatively. Therefore, Model Carving achieves the performance of an overparameterized model at the efficiency of a smaller model.

\begin{table}[t]
\centering
\caption{\textbf{Carved models run up to 23\% fewer inference FLOPs than DenseFormer-style models.} We compare the full-depth model, the DenseFormer-style model from Section \ref{sec:no-removal}, and the carved model. Carved models achieve matched or better performance with fewer inference FLOPs.}
\label{tab:no-removal}
\footnotesize
\setlength{\tabcolsep}{4pt}
\begin{tabular}{@{}llrrrrr@{}}
\toprule
& Model & Params & Tokens & Inference & Val. & CORE-22 \\
& & (M) & (B) & GFLOPs/token & bpb & \\
\midrule
\multicolumn{7}{@{}l}{\emph{Looped Transformer}} \\
370M & full-depth & 375 & 18.8 & 3.039 & 0.808 & 25.7 \\
 & DenseFormer-style & 375 & 18.8 & 3.046 & 0.794 & 27.5 \\
 & carved & 375 & 22.6 & 2.358 & 0.788 & 28.5 \\
\midrule
\multicolumn{7}{@{}l}{\emph{Standard Transformer}} \\
140M & full-depth & 151 & 7.6 & 0.352 & 0.926 & 15.7 \\
 & DenseFormer-style & 151 & 7.5 & 0.353 & 0.920 & 15.2 \\
 & carved & 130 & 8.3 & 0.307 & 0.922 & 16.5 \\
\addlinespace[2pt]
370M & full-depth & 375 & 18.8 & 0.845 & 0.835 & 22.1 \\
 & DenseFormer-style & 375 & 18.8 & 0.845 & 0.831 & 22.1 \\
 & carved & 296 & 22.8 & 0.672 & 0.832 & 22.0 \\
\addlinespace[2pt]
770M & full-depth & 802 & 40.1 & 1.745 & 0.769 & 26.7 \\
 & DenseFormer-style & 802 & 40.1 & 1.746 & 0.762 & 27.4 \\
 & carved & 622 & 50.1 & 1.366 & 0.767 & 27.2 \\
\bottomrule
\end{tabular}
\end{table}

\section{Evaluation protocol}
\label{app:evaluation}

\paragraph{Validation window and bits per byte.}
Every model is evaluated on the same 16{,}384 held-out blocks of 2{,}048 tokens, which hold 151{,}467{,}293 target bytes. We convert each block back to bytes, split into documents at the end-of-text token, and retokenize with the evaluated model's tokenizer. Bits per byte divides the summed loss by the byte count, so we can compare models with different tokenizers.

\paragraph{CORE-22 and the 13 stable tasks.}
CORE-22 \citep{li2024dclm} centers each task's accuracy at its random baseline, $\mathrm{score}_i = (\mathrm{acc}_i - \mathrm{base}_i) / (1 - \mathrm{base}_i)$, and averages the 22 tasks with equal weight. We run lm-evaluation-harness 0.4.12 with few-shot seed 1234 for every model. We obtain our paired intervals for a CORE-22 difference by resampling the items within each task (2{,}000 replicates). The 13 stable tasks are HellaSwag (zero-shot and 10-shot), Jeopardy, BIG-bench QA Wikidata, ARC-Easy, ARC-Challenge, PIQA, OpenBookQA, LAMBADA, Winograd, WinoGrande, SQuAD and CoQA, which, as defined by \cite{heineman2025signal,madaan2024quantifying}, are the tasks whose centered scores increase with model size and has a signal-to-noise ratio of at least 1.5.

\paragraph{Pass@$k$.}
Every prompt starts with the start token, and we perform sampling with top-$p$ 1.0, no top-$k$, and temperature 0.7. Our intervals use 10{,}000 paired bootstrap resamples over problems. When we say ``the full-depth model's inference allowance,'' we mean the largest number of samples the carved model can draw before exceeding those of the full-depth model's $k$ samples: 24 against 16 at standard 2.7B, 22 at standard 1.3B, 20 at looped 1.3B, and 8 at looped 140M or 10 at the other looped sizes against 8. 

Data sources: \texttt{openai/gsm8k} (main), \texttt{HuggingFaceH4/MATH-500}, \texttt{mandarjoshi/trivia\_qa}, \texttt{rajpurkar/squad}, \texttt{EleutherAI/lambada\_openai}.

\paragraph{Decoding throughput.}
After GPU warmup, we time 256 decode steps with random-token prompts at 1{,}024, 4{,}096, and 8{,}192 tokens and batch sizes from 1 to each model's largest tested batch. We test on one H100 80\,GB GPU per pair in two modes: one compiled call per step, and one compiled loop over all 256 steps. Our equal-batch comparisons use four paired repeats in alternating order on the same GPU, and largest-batch rates use four repeats per model. Both models use bfloat16 matrix products, cuDNN prefill attention, and XLA decode attention. 

\begin{figure}[h]
\centering
\includegraphics[width=\textwidth]{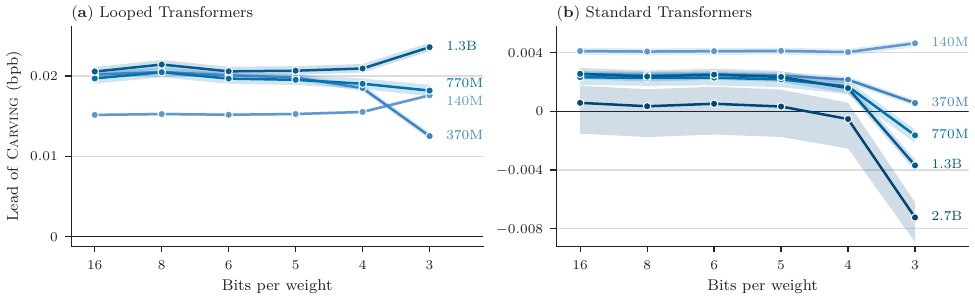}
\caption{\textbf{How robust \carving{} is under weight quantization.} Full-depth bpb minus carved bpb at the same weight precision, for (a) looped and (b) standard Transformers, with 95\% intervals over runs of 64 validation sequences. We trained originally in 16 bits.}
\label{fig:quantization}
\end{figure}

\paragraph{Weight quantization.}
\label{app:quantization}
We quantize the weight matrices of both models of every pair to 8 bits by symmetric rounding per output channel, and to 6, 5, 4 and 3 bits with GPTQ \citep{frantar2023gptq} in groups of 64 columns calibrated on 128 training sequences. Figure~\ref{fig:quantization} shows \carving{}'s lead at each precision (down to 4 bits in every pair except standard 2.7B, and down to 3 bits in the looped family).

\section{Downstream results}
\label{app:downstream}

Table~\ref{tab:core-stable} gives the CORE-22 comparisons on all 22 tasks and on the 13 stable tasks (Appendix~\ref{app:evaluation}), and Table~\ref{tab:passk-looped} gives the pass@$k$ comparisons.

\begin{table}[h]
\centering
\caption{CORE-22 and stable-13 differences with paired 95\% intervals over items (2{,}000 replicates).}
\label{tab:core-stable}
\footnotesize
\begin{tabular}{llrr}
\toprule
Family & Size & CORE-22 & Stable-13 \\
\midrule
Looped & 140M & $+1.06$ [$-0.03$, $+2.13$] & $+1.54$ [$+0.33$, $+2.73$] \\
 & 370M & $+2.75$ [$+1.76$, $+3.79$] & $+4.12$ [$+2.98$, $+5.25$] \\
 & 770M & $+2.33$ [$+1.21$, $+3.48$] & $+3.83$ [$+2.75$, $+4.93$] \\
 & 1.3B & $+0.84$ [$-0.26$, $+1.98$] & $+3.52$ [$+2.54$, $+4.51$] \\
\midrule
Standard & 140M & $+0.76$ [$-0.24$, $+1.81$] & $+1.09$ [$-0.10$, $+2.28$] \\
 & 370M & $-0.04$ [$-1.07$, $+0.95$] & $+0.12$ [$-0.97$, $+1.22$] \\
 & 770M & $+0.45$ [$-0.52$, $+1.46$] & $+0.09$ [$-0.96$, $+1.22$] \\
 & 1.3B & $+1.46$ [$+0.36$, $+2.51$] & $+0.43$ [$-0.56$, $+1.39$] \\
 & 2.7B & $+0.99$ [$-0.07$, $+2.06$] & $+0.35$ [$-0.64$, $+1.33$] \\
\bottomrule
\end{tabular}
\end{table}

\begin{table}[h]
\centering
\caption{Pass@$k$ (temperature 0.7) of the looped pairs within the full-depth model's inference allowance as a fraction of problems, with paired 95\% intervals over problems. }
\label{tab:passk-looped}
\scriptsize
\setlength{\tabcolsep}{2.5pt}
\begin{tabular}{@{}lrrrr@{}}
\toprule
Task & Looped 140M & Looped 370M & Looped 770M & Looped 1.3B \\
Samples (full-depth / carved) & 8 / 8 & 8 / 10 & 8 / 10 & 8 / 10 \\
\midrule
GSM8K & $+$0.011 [$-$0.003,$+$0.025] & \textbf{$+$0.020} [$+$0.004,$+$0.037] & \textbf{$+$0.026} [$+$0.009,$+$0.044] & \textbf{$+$0.066} [$+$0.047,$+$0.084] \\
MATH-500 & $+$0.007 [$-$0.007,$+$0.021] & $+$0.011 [$-$0.005,$+$0.029] & $-$0.012 [$-$0.031,$+$0.006] & \textbf{$+$0.030} [$+$0.009,$+$0.051] \\
TriviaQA & \textbf{$+$0.009} [$+$0.006,$+$0.013] & \textbf{$+$0.035} [$+$0.031,$+$0.039] & \textbf{$+$0.069} [$+$0.065,$+$0.074] & \textbf{$+$0.044} [$+$0.040,$+$0.049] \\
SQuAD & \textbf{$+$0.046} [$+$0.040,$+$0.052] & \textbf{$+$0.056} [$+$0.050,$+$0.062] & \textbf{$+$0.063} [$+$0.058,$+$0.069] & \textbf{$+$0.047} [$+$0.041,$+$0.052] \\
LAMBADA & \textbf{$+$0.019} [$+$0.010,$+$0.028] & \textbf{$+$0.056} [$+$0.047,$+$0.065] & \textbf{$+$0.058} [$+$0.050,$+$0.067] & \textbf{$+$0.052} [$+$0.044,$+$0.060] \\
\bottomrule
\end{tabular}

\end{table}

\paragraph{Released models.}
We compare public, released models in Figure~\ref{fig:released}  and list every model with its CORE-22 score, training compute, and inference cost in Table~\ref{tab:released}. We rescore every released model with public weights with our own harness, requests, and seed (except Parcae, whose baselines have no public weights so we use the published means from \cite{prairie2026parcae}).

\begin{table}[h]
\centering
\caption{CORE-22, training compute, and inference cost of the models in Figure~\ref{fig:released}. ``proxy'' marks an estimate that assumes a 2{,}048-token context. Inference cost is whole-model FLOPs per byte at the average generation context.}
\label{tab:released}
\footnotesize
\setlength{\tabcolsep}{4pt}
\begin{tabular}{@{}lrlrr@{}}
\toprule
Model & Tokens & Training FLOPs & GFLOPs per byte & CORE-22 \\
\midrule
\multicolumn{5}{@{}l}{\emph{Looped Transformers}} \\
\carving{}-140M & 7.9B & $3.1\times10^{19}$ & 0.241 & 19.9 \\
\carving{}-370M & 22.6B & $1.9\times10^{20}$ & 0.522 & 28.5 \\
\carving{}-770M & 52.1B & $8.5\times10^{20}$ & 1.05 & 33.5 \\
\carving{}-1.3B & 82.7B & $2.3\times10^{21}$ & 1.81 & 36.0 \\
Parcae-140M \citep{prairie2026parcae} & 11.2B & $1.5\times10^{19}$ & 0.0887 & 13.5 \\
Parcae-370M & 29.6B & $1.2\times10^{20}$ & 0.275 & 20.0 \\
Parcae-770M & 61.6B & $5.0\times10^{20}$ & 0.607 & 24.1 \\
Parcae-1.3B & 104.0B & $1.6\times10^{21}$ & 1.13 & 29.3 \\
Huginn-3.5B \citep{geiping2025recurrentdepth} & 800.0B & $1.5\times10^{23}$ & 25.1 & 40.7 \\
Nanbeige4.2-3B \citep{nanbeige2026} & 28T & $1.3\times10^{24}$ (proxy) & -- & 58.6 \\
Ouro-1.4B \citep{zhu2025ouro} & 7.7T & $5.1\times10^{23}$ & 2.41 & 54.7 \\
Ouro-2.6B & 7.7T & $7.6\times10^{23}$ & 4.77 & 60.2 \\
\addlinespace[2pt]
\multicolumn{5}{@{}l}{\emph{Standard Transformers}} \\
\carving{}-140M & 8.3B & $9.1\times10^{18}$ & 0.0679 & 16.5 \\
\carving{}-370M & 22.8B & $5.3\times10^{19}$ & 0.149 & 22.0 \\
\carving{}-770M & 50.1B & $2.3\times10^{20}$ & 0.302 & 27.2 \\
\carving{}-1.3B & 89.3B & $5.9\times10^{20}$ & 0.45 & 31.2 \\
\carving{}-2.7B & 196.4B & $2.4\times10^{21}$ & 0.827 & 37.0 \\
DataDecide-750M \citep{magnusson2025datadecide} & 75.0B & $3.5\times10^{20}$ & 0.317 & 27.8 \\
DataDecide-1B & 100.0B & $7.9\times10^{20}$ & 0.538 & 30.5 \\
Delphi-273M \citep{held2026delphi} & 2.2B & $3.0\times10^{18}$ & 0.0789 & 11.3 \\
Delphi-447M & 3.7B & $9.0\times10^{18}$ & 0.143 & 16.0 \\
Delphi-669M ($2\times10^{19}$) & 4.7B & $1.8\times10^{19}$ & 0.228 & 17.3 \\
Delphi-669M ($3\times10^{19}$) & 7.8B & $3.0\times10^{19}$ & 0.228 & 18.2 \\
Delphi-1.4B & 10.6B & $9.0\times10^{19}$ & 0.516 & 23.0 \\
Delphi-1.9B ($2\times10^{20}$) & 14.8B & $1.8\times10^{20}$ & 0.744 & 25.8 \\
Delphi-1.9B ($3\times10^{20}$) & 24.7B & $3.0\times10^{20}$ & 0.744 & 27.7 \\
Delphi-3.4B & 46.3B & $1.0\times10^{21}$ & 1.35 & 33.6 \\
GPT-Neo-125M \citep{black2021gptneo} & 300.0B & $2.6\times10^{20}$ & 0.0595 & 13.2 \\
OPT-125M \citep{zhang2022opt} & 180.0B & $1.8\times10^{20}$ (proxy) & 0.0627 & 13.3 \\
OPT-350M & 180.0B & $4.7\times10^{20}$ (proxy) & 0.166 & 17.1 \\
OPT-1.3B & 180.0B & $1.6\times10^{21}$ (proxy) & 0.616 & 26.2 \\
OPT-2.7B & 180.0B & $3.2\times10^{21}$ (proxy) & 1.23 & 31.3 \\
Pythia-410M \citep{biderman2023pythia} & 299.9B & $8.2\times10^{20}$ & 0.175 & 19.8 \\
Pythia-1B & 299.9B & $1.9\times10^{21}$ & 0.422 & 24.3 \\
Pythia-1.4B & 299.9B & $2.7\times10^{21}$ & 0.611 & 27.0 \\
Llama-3.2-1B \citep{meta2024llama32} & 9T & $7.4\times10^{22}$ (proxy) & 0.551 & 36.6 \\
OLMo-2-1B \citep{olmo2025olmo2} & 4T & $3.8\times10^{22}$ & 0.57 & 41.6 \\
Qwen2.5-0.5B \citep{qwen2025qwen25} & 18T & $6.3\times10^{22}$ (proxy) & 0.233 & 32.0 \\
Qwen2.5-1.5B & 18T & $1.9\times10^{23}$ (proxy) & 0.707 & 46.9 \\
SmolLM2-360M \citep{benallal2025smollm2} & 4T & $1.2\times10^{22}$ & 0.188 & 31.6 \\
SmolLM2-1.7B & 11T & $1.3\times10^{23}$ & 0.802 & 46.7 \\
Parcae baseline-140M \citep{prairie2026parcae} & 11.2B & $5.8\times10^{18}$ & 0.0336 & 13.0 \\
Parcae baseline-370M & 29.6B & $4.2\times10^{19}$ & 0.0914 & 17.5 \\
Parcae baseline-770M & 61.6B & $1.8\times10^{20}$ & 0.193 & 22.4 \\
Parcae baseline-1.3B & 104.0B & $5.5\times10^{20}$ & 0.351 & 25.4 \\
\bottomrule
\end{tabular}
\end{table}

\section{Decoding at every context and batch}
\label{app:decoding}

We report the decoding throughput of the carved model relative to its full-depth model at each context and batch setting in Table~\ref{tab:decode} (see our protocol in Appendix~\ref{app:evaluation}). The ratio grows with the batch, because the carved models move fewer weights and less key--value cache per step (Table~\ref{tab:kv}). On one RTX PRO 6000 workstation GPU with one request at a time, the carved standard 1.3B model decodes 1.31 to 1.34 times faster than its full-depth model and the carved standard 2.7B model 1.42 to 1.44 times faster, at 1{,}024 and 4{,}096 tokens in both timings.

\begin{table}[h]
\centering
\caption{Decoding throughput of the carved model relative to the full-depth model on one H100, by context and batch. Equal batch: the ratio at batch 1 and at the largest batch both models ran (in parentheses), per-step timing. Largest tested batch: each model's largest tested batch before OOM with one compiled call per step and with one compiled loop over the 256 steps. }
\label{tab:decode}
\footnotesize
\setlength{\tabcolsep}{4pt}
\begin{tabular}{llrrrrrr}
\toprule
 & & \multicolumn{2}{c}{Equal batch} & \multicolumn{4}{c}{Largest tested batch} \\
\cmidrule(lr){3-4}\cmidrule(lr){5-8}
Pair & Context & Batch 1 & Largest common & Full-depth & \carving{} & Per step & Loop \\
\midrule
Looped 1.3B & 1,024 & 1.21 & 1.22 (71) & 71 & 83 & 1.26 & 1.31 \\
 & 4,096 & 1.26 & 1.22 (20) & 20 & 24 & 1.29 & 1.28 \\
 & 8,192 & 1.27 & 1.21 (10) & 10 & 12 & 1.29 & 1.37 \\
\addlinespace[2pt]
Standard 1.3B & 1,024 & 1.24 & 1.31 (273) & 273 & 328 & 1.35 & 1.33 \\
 & 4,096 & 1.26 & 1.33 (80) & 80 & 104 & 1.34 & 1.35 \\
 & 8,192 & 1.27 & 1.33 (41) & 41 & 52 & 1.35 & 1.34 \\
\addlinespace[2pt]
Standard 2.7B & 1,024 & 1.34 & 1.43 (163) & 163 & 209 & 1.46 & 1.51 \\
 & 4,096 & 1.39 & 1.45 (48) & 48 & 68 & 1.48 & 1.47 \\
 & 8,192 & 1.42 & 1.45 (24) & 24 & 34 & 1.48 & 1.61 \\
\bottomrule
\end{tabular}

\end{table}

\begin{table}[h]
\centering
\caption{Key--value cache per token. A looped model stores one cache for each loop of each layer. }
\label{tab:kv}
\footnotesize
\begin{tabular}{@{}llrrrr@{}}
\toprule
Family & Size & Attention kept & Full-depth (KiB) & \carving{} (KiB) & Ratio \\
\midrule
Looped & 140M & 61 of 64 & 192.0 & 183.0 & 0.95$\times$ \\
 & 370M & 67 of 80 & 360.0 & 301.5 & 0.84$\times$ \\
 & 770M & 63 of 80 & 540.0 & 425.2 & 0.79$\times$ \\
 & 1.3B & 76 of 96 & 768.0 & 608.0 & 0.79$\times$ \\
\midrule
Standard & 140M & 15 of 16 & 48.0 & 45.0 & 0.94$\times$ \\
 & 370M & 17 of 20 & 90.0 & 76.5 & 0.85$\times$ \\
 & 770M & 17 of 20 & 135.0 & 114.8 & 0.85$\times$ \\
 & 1.3B & 18 of 24 & 192.0 & 144.0 & 0.75$\times$ \\
 & 2.7B & 22 of 32 & 320.0 & 220.0 & 0.69$\times$ \\
\bottomrule
\end{tabular}
\end{table}

\section{Pruning and distillation baselines}
\label{app:baselines}

Table~\ref{tab:pruning} gives the values of Section~\ref{sec:results-pruning} at standard 1.3B. Each pipeline prunes the full-depth model to approximately the inference FLOPs per token as the carved model and retrains it so that the full-depth training and the continued training of the pruned model match the carved run's training FLOPs. The final row prunes the fully trained full-depth model instead and retrains with distillation at 1.21 times the carved run's compute. Table~\ref{tab:distill-intervals} gives the students from Table~\ref{tab:main}.

\begin{table}[h]
\centering
\caption{Pruning the full-depth 1.3B standard Transformer and retraining it, versus the carved model. }
\label{tab:pruning}
\footnotesize
\setlength{\tabcolsep}{4pt}
\begin{tabular}{lrrrr}
\toprule
Model & Compute & Val.\ bpb & CORE-22 & \carving{} $-$ model \\
\midrule
\carving{} & 1.00 & 0.727 & 31.2 & \\
Sheared-LLaMA-style, continued pretraining & 1.00 & 0.742 & 27.6 & $+3.7$ [$+2.5$, $+4.8$] \\
Minitron-style, distillation and cross-entropy & 1.00 & 0.761 & 28.1 & $+3.1$ [$+2.0$, $+4.2$] \\
ShortGPT-style, continued pretraining & 1.00 & 0.745 & 27.9 & $+3.4$ [$+2.3$, $+4.5$] \\
Minitron-style, final full-depth save, distillation only & 1.21 & 0.747 & 29.3 & $+1.9$ [$+0.8$, $+3.1$] \\
\bottomrule
\end{tabular}
\end{table}

\begin{table}[h]
\centering
\caption{Students distilled from the full-depth model of the same size into the shallower-depth shape, against the carved model. Validation bits per byte and CORE-22 of both, and the CORE-22 difference, student minus carved, with its paired 95\% interval over items (2{,}000 replicates). One training run per model.}
\label{tab:distill-intervals}
\footnotesize
\begin{tabular}{lrrrrr}
\toprule
 & \multicolumn{2}{c}{Val.\ bpb} & \multicolumn{2}{c}{CORE-22} & \\
\cmidrule(lr){2-3}\cmidrule(lr){4-5}
Pair & Student & \carving{} & Student & \carving{} & Student $-$ \carving{} \\
\midrule
Standard 140M & 0.947 & 0.922 & 14.23 & 16.51 & $-$2.28 [$-$3.31, $-$1.25] \\
Standard 370M & 0.855 & 0.832 & 19.80 & 22.04 & $-$2.24 [$-$3.22, $-$1.24] \\
Standard 770M & 0.789 & 0.767 & 25.26 & 27.17 & $-$1.91 [$-$2.92, $-$0.89] \\
Standard 1.3B & 0.749 & 0.727 & 29.73 & 31.23 & $-$1.50 [$-$2.51, $-$0.48] \\
Looped 140M & 0.915 & 0.883 & 17.92 & 19.92 & $-$2.00 [$-$3.09, $-$0.87] \\
Looped 370M & 0.828 & 0.788 & 23.81 & 28.48 & $-$4.66 [$-$5.67, $-$3.63] \\
\bottomrule
\end{tabular}

\end{table}

\section{Compute-optimal size and overtraining}
\label{app:compute-optimal}

Table~\ref{tab:isoflop-minima} lists the minima of the fitted curves of Figure~\ref{fig:isoflop}, that is, at each budget, the parameter count and loss at which a quadratic in log parameters through the points of that budget is lowest. The carved minimum is lower at every budget, and the gap grows with the budget.

\begin{table}[h]
\centering
\caption{Compute-optimal size and loss at each budget of Figure~\ref{fig:isoflop} from the fitted curves.}
\label{tab:isoflop-minima}
\footnotesize
\begin{tabular}{llrrrrrr}
\toprule
 & & \multicolumn{2}{c}{Full-depth} & \multicolumn{2}{c}{\carving{}} & & \\
\cmidrule(lr){3-4}\cmidrule(lr){5-6}
Family & Budget (FLOPs) & Params (M) & bpb & Params (M) & bpb & Gap (bpb) & $k^*$ \\
\midrule
Looped & $1.2 \times 10^{19}$ & 187 & 0.931 & 169 & 0.922 & $-$0.009 & 1.16 \\
 & $3.5 \times 10^{19}$ & 441 & 0.868 & 353 & 0.857 & $-$0.011 & 1.25 \\
 & $6.3 \times 10^{19}$ & 579 & 0.840 & 494 & 0.826 & $-$0.014 & 1.31 \\
 & $8.2 \times 10^{19}$ & 688 & 0.826 & 618 & 0.812 & $-$0.014 & 1.32 \\
 & $1.2 \times 10^{20}$ & 951 & 0.807 & 878 & 0.791 & $-$0.016 & -- \\
\midrule
Standard & $1.2 \times 10^{19}$ & 436 & 0.895 & 289 & 0.890 & $-$0.005 & 1.10 \\
 & $3.5 \times 10^{19}$ & 632 & 0.838 & 563 & 0.832 & $-$0.007 & 1.17 \\
 & $6.3 \times 10^{19}$ & 971 & 0.814 & 816 & 0.806 & $-$0.009 & 1.20 \\
 & $8.2 \times 10^{19}$ & 1,073 & 0.802 & 1,043 & 0.792 & $-$0.010 & 1.25 \\
 & $1.2 \times 10^{20}$ & 1,371 & 0.785 & 1,611 & 0.771 & $-$0.014 & -- \\
\bottomrule
\end{tabular}

\end{table}

\paragraph{Overtraining.}
We trained two looped pairs a second time at larger budgets. The 370M pair at $4.9\times10^{20}$ FLOPs, 2.55 times the budget of its main run, gives the full-depth model about 129 tokens per parameter, and it ends at 0.778 bits per byte versus the carved model at 0.758, the same as the main pair's 0.020 with 45 of 160 sublayer applications removed. The 140M pair at $1.2\times10^{20}$ FLOPs (3.8 times the budget of its main run) gives the full-depth model about 191 tokens per parameter, yet it ends at 0.853 bits per byte versus the carved model at 0.836 with 28 of 128 applications removed.

\section{Seeds}
\label{app:seeds}

We trained second carved seeds to evaluate whether \carving{}'s learned topologies and advantage remain stable across seeds (Table~\ref{tab:seeds}). In the standard family the two seeds remove nearly the same sublayers, and in the looped family the three larger pairs share most of the smaller set.

\begin{table}[h]
\centering
\caption{Second carved seeds. Lead: full-depth minus carved validation bpb at equal training compute, with a 95\% block-bootstrap interval over runs of 64 validation sequences, for seed 0 and seed 1. Our count is of closed gates at each completed run's end. }
\label{tab:seeds}
\scriptsize
\setlength{\tabcolsep}{3pt}
\begin{tabular}{llllrl}
\toprule
Pair & Lead, seed 0 & Lead, seed 1 & Closed gates (0 / 1 / shared) & Total & Seed 1 reached \\
\midrule
Standard 140M & 0.0041 [0.0039, 0.0043] & 0.0038 [0.0035, 0.0040] & 5 / 5 / 5 & 32 & the end \\
Standard 370M & 0.0025 [0.0023, 0.0028] & 0.0033 [0.0030, 0.0035] & 9 / 10 / 8 & 40 & the end \\
Standard 1.3B & 0.0026 [0.0022, 0.0030] & 0.0028 [0.0024, 0.0032] & 14 / 13 / 13 & 48 & the end \\
Standard 2.7B & 0.0006 & -- & 22 / 22 / 20 & 64 & update 19{,}600 of 187{,}311 \\
\midrule
Looped 140M & 0.0152 [0.0150, 0.0154] & 0.0156 [0.0155, 0.0158] & 11 / 10 / 4 & 128 & the end \\
Looped 370M & 0.0202 & -- & 36 / 26 / 20 & 160 & update 15{,}210 of 21{,}560 \\
Looped 770M & 0.0197 & -- & 32 / 37 / 24 & 160 & update 6{,}511 of 49{,}730 \\
Looped 1.3B & 0.0206 & -- & 42 / 49 / 39 & 192 & update 25{,}740 of 78{,}840 \\
\bottomrule
\end{tabular}
\end{table}

\section{Instruction tuning}
\label{app:tuning}

Table~\ref{tab:tuned-intervals} shows the paired intervals of the instruct tuned models' differences and the standard 1.3B pair tuned for two epochs. Tuning also increases each model's loss on the pretraining validation window (Table~\ref{tab:forgetting}) by 0.17 to 0.26 bits per byte with the default recipe, and by 0.02 to 0.03 with the gentler recipe.

\begin{table}[h]
\centering
\caption{Tuned models: carved minus full-depth with paired 95\% intervals, on CORE-22 over the 13 stable tasks (centered points) and on chat pass@$k$ (fraction of problems) at equal samples (16 each) and within the nearly equal inference allowance (16 full-depth samples against 24, 22 or 20 carved).}
\label{tab:tuned-intervals}
\scriptsize
\setlength{\tabcolsep}{4pt}
\begin{tabular}{llrrr}
\toprule
Pair, recipe & Measure & Equal samples (16 each) & Nearly equal allowance \\
\midrule
Standard 1.3B, default & CORE-22 (13 stable) & \multicolumn{2}{c}{\textbf{$+$2.78} [$+$1.59, $+$3.93]} \\
 & GSM8K & $-$0.015 [$-$0.041, $+$0.010] & \textbf{$+$0.036} [$+$0.008, $+$0.064] \\
 & TriviaQA & \textbf{$+$0.007} [$+$0.002, $+$0.012] & \textbf{$+$0.033} [$+$0.027, $+$0.038] \\
 & SQuAD & $+$0.005 [$-$0.000, $+$0.011] & \textbf{$+$0.023} [$+$0.018, $+$0.029] \\
\addlinespace[2pt]
Standard 1.3B, gentler & CORE-22 (13 stable) & \multicolumn{2}{c}{$+$0.68 [$-$0.29, $+$1.68]} \\
 & GSM8K & $+$0.014 [$-$0.009, $+$0.037] & \textbf{$+$0.059} [$+$0.034, $+$0.085] \\
 & TriviaQA & $+$0.002 [$-$0.003, $+$0.007] & \textbf{$+$0.026} [$+$0.021, $+$0.032] \\
 & SQuAD & \textbf{$+$0.009} [$+$0.004, $+$0.015] & \textbf{$+$0.026} [$+$0.020, $+$0.031] \\
\addlinespace[2pt]
Standard 2.7B, default & CORE-22 (13 stable) & \multicolumn{2}{c}{\textbf{$+$1.57} [$+$0.57, $+$2.64]} \\
 & GSM8K & $+$0.019 [$-$0.006, $+$0.045] & \textbf{$+$0.073} [$+$0.046, $+$0.101] \\
 & TriviaQA & \textbf{$+$0.023} [$+$0.018, $+$0.028] & \textbf{$+$0.053} [$+$0.047, $+$0.058] \\
 & SQuAD & $-$0.006 [$-$0.010, $-$0.001] & \textbf{$+$0.009} [$+$0.005, $+$0.014] \\
\addlinespace[2pt]
Standard 2.7B, gentler & CORE-22 (13 stable) & \multicolumn{2}{c}{$-$0.45 [$-$1.45, $+$0.54]} \\
 & GSM8K & \textbf{$+$0.027} [$+$0.002, $+$0.051] & \textbf{$+$0.095} [$+$0.067, $+$0.123] \\
 & TriviaQA & $+$0.002 [$-$0.003, $+$0.006] & \textbf{$+$0.028} [$+$0.022, $+$0.033] \\
 & SQuAD & $+$0.002 [$-$0.003, $+$0.007] & \textbf{$+$0.020} [$+$0.015, $+$0.025] \\
\addlinespace[2pt]
Looped 1.3B, default & CORE-22 (13 stable) & \multicolumn{2}{c}{\textbf{$+$5.89} [$+$4.63, $+$7.13]} \\
 & GSM8K & \textbf{$+$0.058} [$+$0.032, $+$0.085] & \textbf{$+$0.086} [$+$0.059, $+$0.114] \\
 & TriviaQA & \textbf{$+$0.055} [$+$0.050, $+$0.060] & \textbf{$+$0.072} [$+$0.066, $+$0.077] \\
 & SQuAD & \textbf{$+$0.007} [$+$0.003, $+$0.012] & \textbf{$+$0.014} [$+$0.010, $+$0.019] \\
\addlinespace[2pt]
Looped 1.3B, gentler & CORE-22 (13 stable) & \multicolumn{2}{c}{\textbf{$+$3.49} [$+$2.43, $+$4.57]} \\
 & GSM8K & \textbf{$+$0.034} [$+$0.009, $+$0.060] & \textbf{$+$0.070} [$+$0.042, $+$0.096] \\
 & TriviaQA & \textbf{$+$0.041} [$+$0.036, $+$0.046] & \textbf{$+$0.057} [$+$0.051, $+$0.062] \\
 & SQuAD & \textbf{$+$0.016} [$+$0.011, $+$0.020] & \textbf{$+$0.024} [$+$0.019, $+$0.029] \\
\addlinespace[2pt]
Standard 1.3B, gentler, two epochs & CORE-22 (13 stable) & \multicolumn{2}{c}{\textbf{$+$1.54} [$+$0.56, $+$2.52]} \\
 & GSM8K & $+$0.010 [$-$0.017, $+$0.036] & \textbf{$+$0.065} [$+$0.036, $+$0.094] \\
 & TriviaQA & $+$0.004 [$-$0.001, $+$0.009] & \textbf{$+$0.028} [$+$0.022, $+$0.033] \\
 & SQuAD & \textbf{$+$0.015} [$+$0.010, $+$0.021] & \textbf{$+$0.031} [$+$0.025, $+$0.037] \\
\bottomrule
\end{tabular}

\end{table}

\begin{table}[h]
\centering
\caption{Forgetting: the rise of each model's loss on the pretraining validation window after tuning in bpb, and the difference, carved minus full-depth (negative: the carved model forgets less).}
\label{tab:forgetting}
\footnotesize
\begin{tabular}{llrrr}
\toprule
Pair & Recipe & Full-depth & \carving{} & Difference \\
\midrule
Standard 1.3B & default & 0.227 & 0.224 & $-0.003$ [$-0.004$, $-0.002$] \\
 & gentler & 0.031 & 0.029 & $-0.002$ \\
Standard 2.7B & default & 0.211 & 0.166 & $-0.045$ [$-0.047$, $-0.043$] \\
 & gentler & 0.022 & 0.022 & $+0.000$ \\
Looped 1.3B & default & 0.263 & 0.253 & $-0.010$ [$-0.010$, $-0.009$] \\
 & gentler & 0.034 & 0.034 & $-0.000$ \\
\bottomrule
\end{tabular}
\end{table}

\section{Cooldown}
\label{app:cooldown}

Table~\ref{tab:cooldown} gives the gain of the final cooldown for every model. The carved model gains 10\% to 16\% less than its full-depth model at every size in both families, and the shallower-depth model gains as much as the full-depth model. We searched the shape of the cooldown, but it does not change the gain.

\begin{table}[h]
\centering
\caption{Gain of the final cooldown, in bpb.}
\label{tab:cooldown}
\footnotesize
\begin{tabular}{llrrrrr}
\toprule
Family & Size & Full-depth & Shallower & Ratio & \carving{} & Ratio \\
\midrule
Looped & 140M & 0.0807 & 0.0811 & 1.00 & 0.0692 & 0.86 \\
 & 370M & 0.0771 & 0.0782 & 1.02 & 0.0653 & 0.85 \\
 & 770M & 0.0739 & 0.0756 & 1.02 & 0.0623 & 0.84 \\
 & 1.3B & 0.0704 & -- & -- & 0.0611 & 0.87 \\
\midrule
Standard & 140M & 0.0734 & 0.0743 & 1.01 & 0.0659 & 0.90 \\
 & 370M & 0.0718 & 0.0719 & 1.00 & 0.0615 & 0.86 \\
 & 770M & 0.0697 & 0.0699 & 1.00 & 0.0612 & 0.88 \\
 & 1.3B & 0.0677 & 0.0690 & 1.02 & 0.0596 & 0.88 \\
\bottomrule
\end{tabular}

\end{table}

\begin{figure}[h]
\centering
\includegraphics[width=\textwidth]{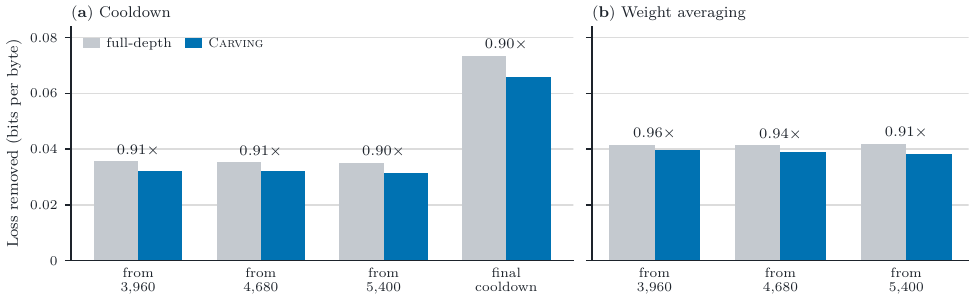}
\caption{\textbf{Cooldown and weight averaging helps the carved model less than non-carved models (140m)} (a) Validation loss decrease from a linear decay of the learning rate to zero (100 updates) at updates 3{,}960, 4{,}680 and 5{,}400, and  the final cooldowns. (b) Loss decrease by averaging weights at the peak learning rate. From each of the checkpoint updates above, we train for 200 more updates at the peak rate and average 20 copies, each 10 updates apart. The ratio above each pair is the carved value divided by full-depth value.}
\label{fig:short-decays}
\end{figure}

We find that \carving{}'s smaller cooldown gain is present at every point of the constant-rate LR phase, not only at the end (Figure~\ref{fig:short-decays}a): 100-update cooldowns from three saves of the standard 140M pair remove 0.90 to 0.91 times as much loss from the carved model as from the full-depth model, and the final cooldowns remove 0.90 times as much. Averaging weights along the constant-rate trajectory also removes less loss from the carved model in the same way (Figure~\ref{fig:short-decays}b).

\section{Synthetic tasks}
\label{app:synthetic}

\paragraph{Tasks.} $S_5$ composition gives a sequence of up to eight permutations of five objects and asks for their composition \citep{liu2022shortcuts}. ListOps \citep{nangia2018listops} gives as input a nested prefix expression over digits with the operators MIN, MAX, median and sum modulo ten, nested at most five deep, and asks for its value. Modular addition takes two numbers below 113 and asks for their sum modulo 113 \citep{nanda2023progress}. Each input ends with a query token, and we apply the loss at that position only.

\paragraph{Model and pressure.} We use a looped Transformer with three layers applied in four loops, width 64 and four attention heads, so 24 sublayer applications. The tasks have no natural training budget so we do not forecast the rate of learning on these tasks. Instead, we use a fixed value $\lambda$ that replaces $v(T)$ in Equation~\eqref{eq:price}, of values 0.28 for $S_5$, 0.4 for ListOps and 0.5 for modular addition. The training budget (for $T$ and $\rho_t$) is the full-depth model's training FLOPs over about 1.05B tokens. Table~\ref{tab:synthetic} contains the final sublayers and accuracy.

\begin{table}[h]
\centering
\caption{Synthetic tasks. The sublayers each carved run keeps, and its exact match accuracy on the held-out test split.}
\label{tab:synthetic}
\footnotesize
\setlength{\tabcolsep}{4pt}
\begin{tabular}{llcp{5.6cm}rr}
\toprule
 & & & & \multicolumn{2}{c}{Test exact match} \\
\cmidrule(lr){5-6}
Task & Seed & Kept & Kept applications & Carved & Full-depth \\
\midrule
Modular addition & 0 & 2 & attention of loop 3 layer 1 and of loop 4 layer 3 & 0.998 & -- \\
 & 1 & 2 & attention of loop 3 layer 1 and of loop 4 layer 2 & 1.000 & -- \\
 & 2 & 2 & attention of loop 3 layer 1 and of loop 4 layer 2 & 0.764 & -- \\
\midrule
ListOps & 0 & 6 & attention of loop 3 layers 1 and 2 and of loop 4 layers 2 and 3; MLP of loop 4 layers 2 and 3 & 0.861 & 0.851 \\
 & 1 & 4 & attention of loop 3 layers 1 and 2 and of loop 4 layer 3; MLP of loop 4 layer 3 & 0.851 & 0.862 \\
 & 2 & 7 & attention of loop 3 layers 1 and 2 and of loop 4 layers 2 and 3; MLP of loop 3 layer 2 and of loop 4 layers 2 and 3 & 0.868 & 0.800 \\
\midrule
$S_5$ composition & 0 & 11 & attention of loop 2 layer 1; attention and MLP of loop 3 layers 1 to 3 and of loop 4 layers 2 and 3 & 1.000 & -- \\
 & 1 & 11 & attention of loop 2 layer 2; attention and MLP of loop 3 layers 1 to 3 and of loop 4 layers 2 and 3 & 1.000 & -- \\
 & 2 & 11 & attention of loop 2 layer 1; attention and MLP of loop 3 layers 1 to 3 and of loop 4 layers 2 and 3 & 1.000 & -- \\
\bottomrule
\end{tabular}
\end{table}

\end{document}